\documentclass{article}

\PassOptionsToPackage{numbers, compress}{natbib}
\usepackage[main, final]{neurips_2026}
\usepackage[utf8]{inputenc} % allow utf-8 input
\usepackage[T1]{fontenc}    % use 8-bit T1 fonts
\usepackage{hyperref}       % hyperlinks
\usepackage{url}            % simple URL typesetting
\usepackage{booktabs}       % professional-quality tables
\usepackage{amsfonts}       % blackboard math symbols
\usepackage{nicefrac}       % compact symbols for 1/2, etc.
\usepackage{microtype}      % microtypography
\usepackage{xcolor}         % colors

\usepackage{amsmath}
\usepackage{multirow}
\usepackage{pifont}
\usepackage[hypcap=true]{caption}
\usepackage[capitalize,noabbrev]{cleveref}
\usepackage[linesnumbered,ruled,vlined]{algorithm2e}
\usepackage{subcaption}
\usepackage{empheq}
\usepackage{graphicx}
\usepackage{enumitem}
\newtheorem{proposition}{Proposition}
\DeclareMathOperator*{\softmax}{softmax}
\usepackage{threeparttable}

\title{Flow-Direct: Feedback-Efficient and Reusable Guidance for Flow Models via Non-Parametric Guidance Field}

\author{
Kim Yong Tan \textsuperscript{\rm 1}~~
Yueming Lyu \textsuperscript{\rm 2,3}\thanks{Corresponding author}~~
Ivor Tsang\textsuperscript{\rm 1,2,3}~~
Yew-Soon Ong\textsuperscript{\rm 1,2,3} \\
\\
\textsuperscript{\rm 1}College of Computing and Data Science, Nanyang Technological University, Singapore\\
\textsuperscript{\rm 2}Centre for Frontier AI Research, Agency for Science, Technology and Research, Singapore\\
\textsuperscript{\rm 3}Institute of High Performance Computing,  Agency for Science, Technology and Research, Singapore\\
\\
{\tt \{KIMYONG001,ASYSOng\}@ntu.edu.sg}\\
{\tt \{Lyu\_Yueming,Ivor\_Tsang\}@cfar.a-star.edu.sg}
}

\begin{document}

\maketitle

\begin{abstract}
Training-free guidance enables pre-trained diffusion and flow models to optimize application-specific objectives using feedback from external black-box reward functions. However, existing methods are feedback-inefficient because reward feedback is used only transiently to inform a localized gradient approximation or a discrete search decision, and is subsequently discarded. To address this limitation, we propose Flow-Direct, a framework that guides the generation process via a persistent \textit{guidance field}. Theoretically, this guidance field is analytically derived from the log-density ratio between the base and reward-weighted target distributions; it transports the pre-trained distribution to the target distribution. In practice, the field is implemented as a non-parametric estimator constructed from all accumulated reward-evaluated samples. As more samples are collected during optimization, this empirical guidance field becomes increasingly accurate. This persistent formulation yields two major advantages. First, Flow-Direct is highly \textit{feedback-efficient}: because every evaluated sample is used to refine the global guidance field, no reward information is wasted. Second, the framework is naturally \textit{reusable}: once optimization is complete, the collected dataset defines a reusable guidance field for generating novel target samples without additional reward evaluations, and distinct guidance fields can be combined to generate samples that simultaneously satisfy multiple objectives. The source code is publicly available at: \url{https://github.com/kimyong95/finetune-stable-diffusion}.
\end{abstract}

\section{Introduction}

Diffusion and flow-based generative models have become a dominant paradigm for high-dimensional generation. They achieve strong performance in image synthesis~\citep{rombach2022high,esser2024scaling}, video generation~\citep{polyak2024movie,liu2024sora}, and audio synthesis~\citep{le2023voicebox}, and have increasingly been adopted in scientific domains such as molecule generation~\citep{hoogeboom2022equivariant,guan20233d}, protein design~\citep{watson2023novo}, and materials discovery~\citep{zeni2023mattergen}. This success stems from their ability to model complex, multi-modal data distributions and to transform simple noise into realistic samples through a learned generative trajectory.

However, many applications do not only require samples from the broad distribution learned during pre-training. Instead, they require samples from a specific target distribution defined by application-specific objectives. For example, protein design may seek structures with high binding affinity or stability, while aerodynamic design may seek shapes with low drag, high efficiency, or other physical properties. In such settings, the target distribution is often specified only implicitly through an external reward function that evaluates the performance of a generated sample. This reward evaluation process may be a laboratory experiment, a computational simulation, or a learned preference model. Because these evaluation procedures are often expensive, non-differentiable, and unavailable during model training, target generation must be achieved with limited black-box feedback.

A prominent approach for this setting is \emph{training-free guidance}, which steers a pre-trained generative model at inference time using feedback from an external reward function. Since these methods do not update model parameters, they avoid the cost and instability of fine-tuning or reinforcement learning, and can be applied when the reward function is black-box or non-differentiable.

To achieve this, recent methods typically rely on gradient approximation, sequential Monte Carlo (SMC), or tree search. Specifically, during each intermediate denoising step, Evolvable~\citep{wei2025evolvable} guides the generation trajectory via zero-order gradient estimation, and FK-Steering~\citep{singhal2025general} performs particle resampling (SMC) to steer towards the target distribution. On the other hand, TreeG~\citep{guo2025training} formulates the sequential generation process as a search tree problem, and DSearch~\citep{li2025dynamic} incorporates dynamic tree expansion for efficient traversal. Despite their flexibility, they remain feedback-inefficient, requiring a massive number of reward evaluations for accurate gradient approximation, robust particle resampling, or comprehensive search. We defer a more detailed review to~\cref{app:related_works}.

We argue that a fundamental source of this inefficiency is that existing inference-time guidance methods use feedback only transiently. At each generation step, samples are evaluated to estimate a local gradient, select a search branch, or approximate a posterior distribution. Once this intermediate decision is made, the evaluated samples are discarded. As a result, valuable information from expensive reward evaluations is not preserved, leading to low feedback efficiency.

To address this limitation, we propose Flow-Direct, a framework that steers the flow models via a persistent \textit{guidance field}. Theoretically, the guidance field is analytically derived from the log-density ratio between the base and reward-weighted target distributions; it transports the pre-trained distribution to the target distribution. In practice, the guidance field is implemented as a non-parametric estimator constructed from the reward-evaluated samples.

Flow-Direct is highly \textit{feedback-efficient} because rather than utilizing reward feedback transiently, the non-parametric implementation allows us to aggregate the evaluated samples persistently, ensuring no reward-information loss. Over the course of the optimization process, all reward-evaluated samples are appended to a growing empirical dataset. As this dataset expands, the non-parametric estimator yields an increasingly accurate approximation of the exact guidance field, thereby guiding the flow model towards the target distribution effectively.

Furthermore, this framework is naturally \textit{reusable}. Once a labeled  dataset is collected for a specific target objective, its corresponding empirical guidance field can be reused to generate novel target samples without further reward evaluations. Additionally, the non-parametric estimators learned from distinct objectives can be linearly combined to produce samples that simultaneously satisfy multiple targets.

Our contributions are summarized as follows:
\begin{itemize}[topsep=0pt, leftmargin=2.5em, itemsep=0pt]
\item We propose Flow-Direct, a training-free framework that guides pre-trained flow models using a \textit{non-parametric guidance field}. This field is analytically derived from the log-density ratio between the base and reward-weighted target distributions, and is implemented as a non-parametric estimator.
\item Moreover, Flow-Direct is highly \textit{feedback-efficient}. Because the non-parametric implementation utilizes accumulated reward-evaluated samples rather than using feedback transiently, no reward information is discarded. The reward feedback is persistently preserved in the guidance field, which becomes increasingly accurate throughout the optimization process.
\item Furthermore, the constructed guidance field is natively \textit{reusable} after optimization. Once a labeled dataset is collected, it can steer the flow model to generate novel target samples without requiring further reward evaluations. Moreover, multiple empirical fields can be linearly combined to generate samples that simultaneously satisfy multiple objectives.
\item Finally, we conduct comprehensive experiments on both image and 3D generation tasks, indicating Flow-Direct's superior effectiveness, scalability, and robust performance across multiple domains compared to existing methods.
\end{itemize}

\section{Preliminaries}

\textbf{Flow Model.} The flow matching framework~\citep{lipman2022flow}
constructs a continuous-time transport from a noise distribution
$p_0(x_0) = \mathcal{N}(0,I)$ to the data distribution $p_1(x_1)$ via
linear interpolation $x_t = (1-t)\,x_0 + t\,x_1$ for $t \in [0,1]$.
Conditioned on a data point $x_1$, the intermediate state admits a
Gaussian distribution:
\begin{align}
p_t(x_t | x_1) = \mathcal{N} \big(x_t \big| t x_1,  (1-t)^2 I\big).
\label{eq:conditional}
\end{align}
The trajectory of $x_t$ is governed by the ODE $dx_t = v_t(x_t)\, dt$,
with the time-dependent vector field given as:
\begin{align}
v_t(x_t) = \frac{x_t}{t} + \frac{1-t}{t} \nabla \log p_t(x_t),
\label{eq:vector_score}
\end{align}
where we provide the derivation in~\cref{app:proofs}. In practice, a neural network $v_\theta(x_t)$ is trained to approximate
the true velocity field, i.e., $v_\theta(x_t) \approx v_t(x_t)$. During
inference, new samples are generated by drawing $x_0 \sim \mathcal{N}(0,I)$
and integrating the ODE. A standard and simple choice for solving the ODE is
first-order Euler discretization:
\begin{align}
    x_{t+\Delta t}
    = \underbrace{x_t + \Delta t\, v_t}_{\texttt{step}(x_t, v_t)},
\label{eq:ode_step}
\end{align}
where $v_t \coloneqq v_\theta(x_t)$. Iterating this update from $t=0$ to
$t=1$ produces a final sample $x_1 \sim p_1^\theta(x_1)$ from the learned
data distribution. In this work, we treat the numerical update operator
\texttt{step} as a black-box sampler, which may be instantiated by more
advanced ODE or SDE solvers, and we focus solely on modifying the vector field
$v_t$.

\section{Method}
In this section, we introduce Flow-Direct, a training-free framework that steers pre-trained flow models via a \textit{non-parametric guidance field}. We begin by formulating this guidance field in~\cref{sec:3.1}, showing how it can be constructed when target data is directly available. Since such target data is typically unavailable in real-world scenarios, \cref{sec:3.2} presents an optimization algorithm that estimates the guidance field using an external black-box reward. This optimization process naturally produces a labeled target dataset. In~\cref{sec:3.3}, we show how the accumulated dataset can be reused to generate novel target samples without additional reward evaluations. Finally,~\cref{sec:3.4} highlights the theoretical properties and practical advantages of our framework.

\subsection{Constructing the Guidance Field from Target Data}
\label{sec:3.1}

\textbf{Problem Setup.} Given a pre-trained flow model, we consider the problem of steering its generative distribution toward a desired target distribution using only two small datasets, without retraining. Formally, we assume access to a base dataset, sampled from a distribution that approximates the pre-trained distribution; and a target dataset, sampled from the desired distribution. Our goal is to construct a guidance formulation, computed entirely from these two datasets, that steers the pre-trained distribution from the base toward the target.

\begin{figure}[ht]
  \centering
  \includegraphics[width=0.8\linewidth]{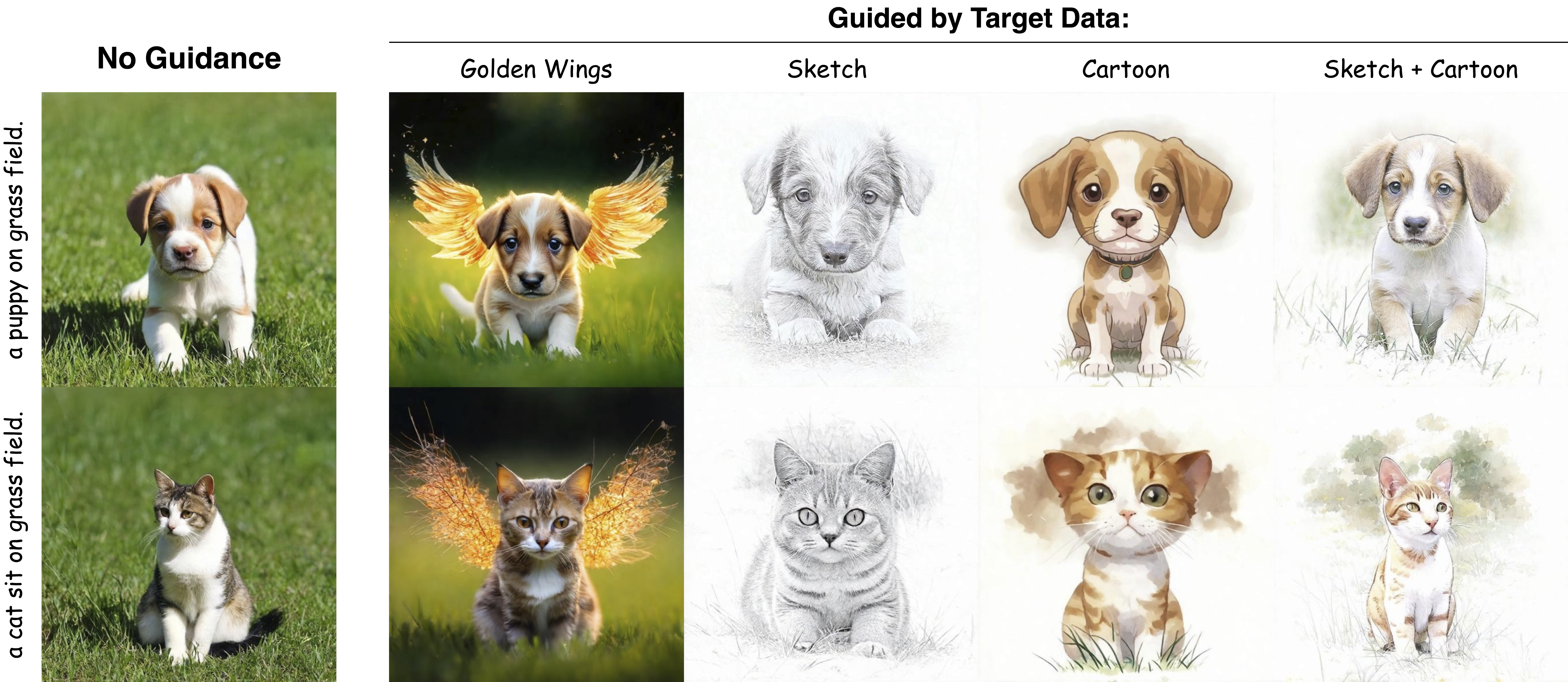}
\caption{\textbf{Guided Flow Model by Target Data.}
Each row shows generations from the same prompt: the first column is the unguided baseline, and the remaining columns apply~\cref{eq:data_field} with different target datasets. The \textit{Sketch + Cartoon} column is produced by summing the two shift terms.
\textbf{Top:} guidance steers the \textit{puppy} generation toward each target distribution, and the composed shift meaningfully mixes both styles.
\textbf{Bottom:} even when the prompt (\textit{cat}) mismatches the guidance datasets (\textit{puppy}), the target attributes still transfer. This shows that our formulation generalizes beyond the matched-distribution setting.}
\label{fig:data_guidance}
\end{figure}

\textbf{Guided Score.}
Let $v_\theta$ denote a pre-trained generative flow model, and let $p_t^\theta$ denote the marginal distribution that this model generates at time $t$. Suppose the flow model is well-trained to approximate the true vector field, i.e., $v_\theta(x_t) \approx v_t(x_t)$ in~\cref{eq:vector_score}, then:
\begin{align}
v_\theta(x_t) \approx \frac{x_t}{t} + \frac{1-t}{t} \nabla \log p_t^\theta(x_t).
\label{eq:vector_score_theta}
\end{align}
This identity shows that
the learned flow implicitly encodes the score of $p_t^\theta$, which allows us to perform guidance in score space.

Let $p_1^\text{base}$ and $p_1^\text{target}$ denote the base and target data distributions. For each, we use~\cref{eq:conditional} to construct a corresponding probability path, $p_t^\text{base}$ and $p_t^\text{target}$ for $t \in [0,1]$, by applying the flow matching interpolation between noise and the respective endpoint distribution. We guide the flow model in the score space by adding the log-density ratio between the base and target distribution:
\begin{align}
v^\text{guided}_\theta(x_t) & \coloneqq \frac{x_t}{t} + \frac{1-t}{t} \left(\nabla \log p_t^\theta(x_t) + \nabla_{x_t} \log \frac{p^\text{target}_t(x_t)}{p^\text{base}_t(x_t)} \right)
\label{eq:guided_flow_1} \\
& = v_\theta(x_t) + \underbrace{\frac{1}{1-t}
    \left(\mathbb{E}_{p_1^{\textnormal{target}}}[x_1 | x_t] - \mathbb{E}_{p_1^{\textnormal{base}}}[x_1 | x_t]\right)
}_{\textnormal{guidance field }\Delta(x_t)},
\label{eq:guided_flow_2}
\end{align}
and we show that this guidance is equivalent to a guidance field $\Delta(x_t)$ constructed by the difference between the base and target posterior expectation, with the derivation provided in~\cref{app:proofs}. In~\cref{eq:guided_flow_1}, the base distribution has a role of reference distribution, suppose $p^\text{base}_t(x_t) = p^\theta_t(x_t)$, then they cancel out, and the guided flow model generates the target distribution exactly. In~\cref{eq:guided_flow_2}, the posterior expectation $\mathbb{E}[x_1|x_t]$ represents the ideal estimate of the clean data $x_1$ given the current state $x_t$ at time $t$. Geometrically, it is a vector pointing from the base data towards the target data. In practice, we implement the guidance field by approximating the posterior expectation using the non-parametric estimator given in~\cref{proposition:1} with the derivation provided in~\cref{app:proofs}:

\begin{proposition}[Estimator for Posterior Expectation]
\label{proposition:1}
Let $\mathcal{D}=\{x_1^i\}_{i=1}^{N}$ be a dataset where $x_1^i \overset{\textnormal{iid}}{\sim} p_1^\textnormal{base}(x_1)$. Under the flow matching framework, for any $x_t$
with $t \in [0,1)$, the posterior expectation can be estimated by:
\begin{align}
\sum_{i=1}^N
\underbrace{
  \softmax_{i' \in [N]}\!\left(-\frac{\|x_t-t x_1^{i'}\|^2 }{2(1-t)^2}\right)_{\!\!i}
}_{\hat{p}_1(x_1=x_1^i | x_t)} x_1^i
\;\xrightarrow{N\to\infty}\;
\mathbb{E}_{p_1}[x_1 | x_t],
\label{eq:posterior_estimator}
\end{align}
and the estimator converges to the true posterior expectation as the dataset size grows to infinity.
\end{proposition}

Suppose we have access to the base and target datasets, $\mathcal{D}_\text{base} = \{ x_1^i \}_{i=1}^N$ where $x_1^i \overset{\text{iid}}{\sim} p_1^\text{base}(x_1)$ and $\mathcal{D}_\text{target} = \{ x_1^j \}_{j=1}^M$ where $x_1^j \overset{\text{iid}}{\sim} p_1^\text{target}(x_1)$. Applying~\cref{proposition:1} yields a non-parametric guidance field:\footnote{See~\cref{eq:data_field_practical} for the corresponding practical implementation.}

\begin{equation}
\label{eq:data_field}
\resizebox{\linewidth}{!}{$\displaystyle
\boxed{
\Delta(x_t;\mathcal{D}_\text{base},\mathcal{D}_\text{target}) \coloneqq \frac{1}{1-t} \Bigg[ \underbrace{
\sum_{j=1}^M \softmax_{j' \in [M]} \left( -\frac{\|x_t - t x_1^{j'}\|^2}{2(1-t)^2} \right)_{\!\!j} \! x_1^j}_{\displaystyle \approx \mathbb{E}_{p^\text{target}_1}[x_1 | x_t] } - \underbrace{\sum_{i=1}^N \softmax_{i' \in [N]} \left( -\frac{\|x_t - t x_1^{i'}\|^2}{2(1-t)^2} \right)_{\!\!i} \! x_1^i }_{\displaystyle \approx \mathbb{E}_{p^\text{base}_1}[x_1 | x_t] } \Bigg] \!
}
$}
\end{equation}

where we use the shorthand notation for softmax\footnote{We use the shorthand notation: $\operatorname{softmax}_{i' \in [N]}(z_{i'})_i \coloneqq \exp(z_i) / \sum_{i'=1}^N \exp(z_{i'})$ where $i'$ is a dummy index ranging over $[N]=\{1,\cdots,N\}$, and the outer subscript $i$ selects the $i$-th component of the resulting probability vector.} throughout this paper. Substituting this guidance field into~\cref{eq:guided_flow_2} yields the guided flow model. While this provides the exact theoretical formulation, direct computation in high-dimensional spaces can suffer from numerical instability. We discuss the practical implementation details in~\cref{app:implementation}.

\textbf{Demonstration.} We test the non-parametric guidance field formulation. Our goal is to steer the pre-trained flow model from the \textit{puppy} base distribution toward three distinct targets: \textit{puppy with golden wings}, \textit{sketch drawing of puppy}, and \textit{cartoon puppy}. For each target, we prepare a pair of datasets, $\mathcal{D}_\text{base}$ and $\mathcal{D}_\text{target}$, consisting of 64 images each (details provided in~\cref{app:demo_impl}), and construct the corresponding guidance field according to~\cref{eq:data_field}. As shown in~\cref{fig:data_guidance}, applying the guidance field successfully steers the flow model toward each respective target distribution. Furthermore, linearly combining the \textit{sketch} and \textit{cartoon} guidance fields yields meaningful compositional generation. We also evaluate a distribution-mismatch setting ($p^\theta_t \neq p^\text{base}_t$) by using the inference prompt \textit{cat} while applying the guidance field derived from the \textit{puppy} datasets. The target attributes are successfully transferred, demonstrating that our formulation remains robust even when the base dataset does not match the pre-trained distribution.

\subsection{Optimizing the Guidance Field via Black-Box Rewards}
\label{sec:3.2}
\textbf{Problem Setup.} In practical settings, the target dataset is not directly accessible. In this section, we discuss how to learn the guidance field using an external reward function. Suppose we have access to a base data distribution $p^\text{base}_1(x_1)$ and an external reward function $r(x_1)$. Our goal is to steer generation toward a target distribution biased toward high-reward regions. We define the target distribution as a Boltzmann distribution:
\begin{align}
p^\text{target}_1(x_1) \propto p^\text{base}_1(x_1)\exp(r(x_1)).
\label{eq:target_distribution}
\end{align}
Following~\cref{eq:guided_flow_2}, the guidance field toward this target distribution is determined by the target posterior expectation. We show that this expectation admits a non-parametric estimator in~\cref{proposition:2} with the derivation provided in~\cref{app:proofs}:
\begin{proposition}[Estimator for Target Posterior Expectation]
\label{proposition:2}
For the target distribution $p_1^\textnormal{target}(x_1)$ and reward function $r(x_1)$ defined in~\cref{eq:target_distribution}, let $\mathcal{D}=\{(x_1^i,r_i)\}_{i=1}^{N}$ be a labeled dataset, where $x_1^i \overset{\textnormal{iid}}{\sim} p_1^\textnormal{base}(x_1)$ and $r_i=r(x_1^i)$. Under the flow matching framework, for any $x_t$ with $t \in [0,1)$, the target posterior expectation can be estimated by:
\begin{align}
\sum_{i=1}^N
\softmax_{i' \in [N]}\!\left(-\frac{\|x_t-tx^{i'}_1\|^2}{2(1-t)^2} + r_{i'}\right)_{\!\!i} x_1^i
\;\xrightarrow{N\to\infty}\;
\mathbb{E}_{p^\textnormal{target}_1}[x_1 | x_t],
\label{eq:target_posterior_estimator}
\end{align}
which converges to the true target posterior expectation as the dataset
size grows to infinity.
\end{proposition}
Compared to~\cref{proposition:1}, each softmax logit has the corresponding reward $r_{i'}$ added to it, biasing the distribution toward high-reward samples and thereby biasing the estimator toward the target distribution. The guidance field to transport $p_1^\text{base}(x_1)$ to $p_1^\text{target}(x_1)$ is thereby given as:\footnote{See~\cref{eq:reward_guidance_practical} for the corresponding practical implementation.}
\begin{align}
\boxed{
\Delta(x_t;\mathcal{D}) \coloneqq \frac{1}{1-t}  \sum_{i=1}^N \left[ \; \softmax_{i'\in [N]}\left(-\frac{\|x_t - t x_1^{i'}\|^2}{2(1-t)^2 } + r_{i'}\right)_{\!\!i} - \softmax_{i'\in [N]} \left(-\frac{\|x_t - t x_1^{i'} \|^2}{2(1-t)^2 }\right)_{\!\!i} \; \right] x^i_1
}
\label{eq:reward_guidance}
\end{align}

\textbf{Iterative Optimization.} A single application of~\cref{eq:reward_guidance} produces a target distribution that scales the base by a factor of $\exp(r(x_1))$. To push further toward high-reward regions, we iteratively define a sequence of distributions, where the target of the current iteration becomes the base for the next. Specifically, starting from the pre-trained distribution $p^0_1 \coloneqq p^\text{base}_1$, we define a sequence of distributions recursively as:
\begin{align}
p^l_1(x_1) \propto p^{l-1}_1(x_1)\exp(r(x_1)) &\propto p^{0}_1(x_1)\exp(l \cdot r(x_1)),
\label{eq:target_distribution_iterative}
\end{align}
where the second proportional follows by unrolling the recursion. At each $l$-th iteration, the guidance field that transports $p^{0}_1(x_1)$ to $p^l_1(x_1)$ is therefore:
\begin{align}
\frac{1}{1-t} \left(\mathbb{E}_{p_1^l}[x_1 | x_t] - \mathbb{E}_{p_1^0}[x_1 | x_t] \right) 
= \sum_{0\le j < l} \frac{1}{1-t} \left(\mathbb{E}_{p_1^{j+1}}[x_1 | x_t] - \mathbb{E}_{p_1^j}[x_1 | x_t] \right)
\approx \sum_{0\le j<l} \Delta(x_t;\mathcal{D}_j),
\label{eq:telescope}
\end{align}
where the first equality decomposes into a telescoping sum of per-step guidance fields, and the second approximation follows from the per-step non-parametric guidance field in~\cref{eq:reward_guidance}. After completing the $l$-th iteration, we obtain the new target dataset $\mathcal{D}_l \sim p^l_1(x_1)$. This process repeats. We summarize the full procedure in Algorithm~\ref{alg:main}.

\begin{algorithm}[t]
  \DontPrintSemicolon
  \caption{Flow-Direct}
  \label{alg:main}

  \KwIn{Pre-trained flow model $v_\theta$, reward function $r(\cdot)$, maximum optimization iterations $L$, batch size $N$, step size $\Delta t$, guidance field $\Delta(x_t;\mathcal{D})$ as defined in~\cref{eq:reward_guidance_practical}.}
  \KwOut{Generated samples $\{x_1\}$ and accumulated labeled datasets $\mathcal{D}_\text{all} = \bigcup_{l=0}^{L-1} \mathcal{D}_l$.}

  \SetKwBlock{DoParallel}{do parallel for $N$ instances}{end}

  \For{$l \gets 0$ \KwTo $L-1$}{
  \DoParallel{
    Sample $x_0 \sim \mathcal{N}(0, I)$
    
    \For{$t \in [0,1)$ \textnormal{with step} $\Delta t$}{
      
      $v_t \gets v_\theta(x_t, t) + \sum_{0\le j<l} \Delta(x_t;\mathcal{D}_j)$
      
      $x_{t+\Delta t} \gets \texttt{step}(x_t,v_t)$
    }

    Evaluate reward $r \gets r(x_1)$
  }
  Collect dataset $\mathcal{D}_l \gets \{ (x^i_1, r_i ) \}_{i=1}^N$
  }
\end{algorithm}

\subsection{Reusable Guidance Fields}
\label{sec:3.3}

Over the course of optimization, we aggregate the labeled datasets across all $L$ iterations:
\begin{align}
	\mathcal{D}_\text{all} \leftarrow \mathcal{D}_0 \cup \mathcal{D}_1 \cup \cdots \cup \mathcal{D}_{L-1} = \{(x^i_1,r_i)\}_{i=1}^{NL}.
\end{align}
and we denote the underlying distribution as $x^i_1 \sim q(x_1)$. Since it contains samples drawn from a sequence of distributions ranging from the initial base to the final target, its distribution lies between them. We therefore define the contrastive distributions symmetrically around $q(x_1)$:
\begin{equation}
\left\{
\begin{aligned}
    & \; p^+(x_1) \propto q(x_1) \exp(\alpha r(x_1)) \\
    & \; q(x_1) \propto p^-(x_1) \exp(\alpha r(x_1))
\end{aligned}
\right.
\end{equation}
where $\alpha \ge 0$ controls the gap between the $p^+(x_1)$ and $p^-(x_1)$. By rearranging $p^-(x_1)$ and applying~\cref{proposition:2} to both distributions, the guidance field to transport $p^-(x_1)$ to $p^+(x_1)$ is:\footnote{See~\cref{eq:reuse_guidance_practical} for the corresponding practical implementation.}
\begin{equation}
\label{eq:reuse_guidance}
\resizebox{\linewidth}{!}{$\displaystyle
\boxed{
\Delta(x_t,\mathcal{D}_\text{all}, \alpha) \coloneqq \frac{1}{1-t}  \sum_{i=1}^N \left[ \softmax_{i' \in [N]}\left(-\frac{\|x_t - t x_1^{i'}\|^2}{2(1-t)^2 } + \alpha r_{i'} \right)_{\!\!i } - \softmax_{i' \in [N]} \left(-\frac{\|x_t - t x_1^{i'} \|^2}{2(1-t)^2} - \alpha r_{i'} \right)_{\!\! i \; } \right] x^i_1
}
$}
\end{equation}
where $\alpha$ also controls the strength of the guidance signal. Furthermore, given a collection of $K$ datasets $\{\mathcal{D}_\text{all}^k\}_{k=1}^K$ optimized for distinct rewards, we can naturally extend this formulation to achieve compositional multi-target generation. Specifically, we construct a mixed guidance field by linearly combining the guidance fields associated with the $K$ targets:
\begin{align}
    \Delta_{\text{mix}}(x_t) = \sum_{k=1}^K \Delta(x_t; \mathcal{D}_\text{all}^k, \alpha_k),
\label{eq:reuse_multi}
\end{align}
where each $\alpha_k$ controls the guidance strength and trade-off of each $k$-th objective. This mixture of guidance field guides the flow model to generate samples that simultaneously satisfy multiple target rewards, without requiring further reward evaluations or multi-objective optimization.

\subsection{Properties and Advantages}
\label{sec:3.4}

In this section, we highlight several properties and advantages of Flow-Direct.

\textbf{Regularized Guidance.} The guidance direction induced by the non-parametric field is a linear combination of empirical data points. Consequently, the guidance vector is confined to the linear subspace spanned by the dataset. Under the data manifold hypothesis, the guidance field is approximately tangent to the underlying data manifold, which helps regularize the guided distribution and keeps it close to the data manifold, thereby mitigating the artifacts and degeneration commonly observed in unconstrained guidance methods.

\textbf{Sampler-Agnostic.} Flow-Direct directly modifies the continuous vector field, thus it is fundamentally agnostic to the choice of numerical sampler. This contrasts with SMC-based or search-based methods, which rely on specific SDE processes for stochastic exploration, and their performance can be sensitive to sampler configurations and noise schedules. As a result, our framework ensures stable performance and broader applicability across different sampler implementations.

\textbf{Tuning-Free.} Flow-Direct (Algorithm~\ref{alg:main}) relies on a closed-form derivation of the guidance field. In contrast to heuristic search or gradient-based methods, it does not introduce complex, sensitive parameters such as learning rates, momentum, or step sizes. Aside from the reward evaluation budget (batch size $N$ and optimization iterations $L$), the formulation requires no optimization-specific tuning, significantly reducing implementation complexity.

\section{Experiments}

\subsection{Target Distribution Optimization}
In this section, we evaluate Flow-Direct (Algorithm \ref{alg:main}) against baseline methods on image alignment tasks and 3D optimization task, using a matched reward-evaluation budget.

\textbf{Pre-trained Flow Models.} For the \underline{image task}, we use Stable Diffusion 3~\citep{esser2024scaling} (\texttt{stable-diffusion-3.5-medium}) as our text-to-image model; for the \underline{3D task}, we use TRELLIS~\citep{xiang2025structured} (\texttt{microsoft/TRELLIS-text-xlarge}) as our text-to-3D model. For both models, we adopt identical sampling parameters: sampling steps $T=50$, classifier-free guidance scale $\texttt{cfg}=4.5$, and the SDE sampler described in~\cref{eq:sde_step} with noise scale $\eta=0.7$.

\textbf{Reward Functions.} For the \underline{image task}, we evaluate against six external reward functions. \textit{Aesthetic}~\citep{schuhmann2022laion} and \textit{HPSv3}~\citep{ma2025hpsv3} are pre-trained predictors that estimate perceptual image quality and human preference, respectively. \textit{Compressibility} and \textit{Incompressibility}~\citep{black2023training} reward small and large JPEG-compressed file size, respectively. For \textit{Attribute Alignment}, we use open source VLM Gemma~\citep{team2024gemma} to evaluate whether the generated image matches a specified semantic attribute of the animal prompt; for \textit{Prompt Alignment}, we use the VLM Gemini~\citep{reid2024gemini,comanici2025gemini} through paid API to evaluate the alignment with a specified target prompt. The implementation details are deferred to~\cref{app:attribute_alignment} and~\cref{app:prompt_alignment}.

For the \underline{3D task}, we consider vehicle aerodynamic optimization  by minimizing the drag coefficient. Specifically, we prompt the model with \textit{car} to generate a 3D car model, and utilize the DoMINO~\citep{ranade2025domino} to evaluate its drag coefficient, with implementation details provided in \cref{app:domino}.

We consider reward evaluations to be expensive and impose a fixed budget of \underline{$800$} reward evaluations per run for the \textit{Prompt Alignment} task, and \underline{$1600$} for all other tasks.

\textbf{Experimental Procedure.} We implement Flow-Direct (Algorithm~\ref{alg:main}) with batch size $N=16$ for all tasks, and set the number of optimization iterations to $L=50$ for the \textit{Prompt Alignment} task and $L=100$ for all other tasks, so that $N \cdot L$ matches the given reward evaluation budget. We compare Flow-Direct against four inference-time baseline methods: Evolvable~\citep{wei2025evolvable}, FK-Steering~\citep{singhal2025general}, TreeG~\citep{guo2025training}, and DSearch~\citep{li2025dynamic}. For each, we adjust batch-size-related hyperparameters to match the reward evaluations budget and keep all remaining hyperparameters at their officially recommended values. We use the same hyperparameters for all methods for both image and 3D tasks.

For \textit{Aesthetic}, \textit{HPSv3}, \textit{Compressibility}, and \textit{Incompressibility}, we run each method on six simple animal prompts; for \textit{Attribute Alignment}, we assign each animal with a semantic target attribute and repeat each over three seeds, for \textit{Prompt Alignment}, we use the complex prompt and repeat each over three seeds. The specific prompts are stated in~\cref{tab:attribute_alignment_pairs} and~\cref{app:prompt_alignment}. For the aerodynamic optimization, we repeat each experiment over three seeds.

\begin{table}[ht]
  \caption{Reward value (higher is better) achieved by each method across all reward functions. For each run we record the average reward over the final batch of generated samples, and report the mean and standard deviation across runs. Note: As \textit{Compressibility} and \textit{Aerodynamic} are minimization tasks, their values are inverted.}
  \label{tab:result}
  \resizebox{\textwidth}{!}{
  \begin{tabular}{lcccccccccccc}
    \toprule
    \multirow{2}{*}{Algorithm} & \multirow{2}{*}{Aesthetic} & \multirow{2}{*}{HPSv3} & \multirow{2}{*}{Compressibility} & \multirow{2}{*}{Incompressibility} & \multicolumn{6}{c}{Attribute Alignment} & \multirow{2}{*}{Prompt Alignment} & \multirow{2}{*}{Aerodynamic} \\
    \cmidrule(lr){6-11}
     & & & & & Happy & Fluffy & Running & Cute & Vivid & Fierce & & \\
    \midrule
    Evolvable    & $6.37{\pm}.54$ & $8.57{\pm}.96$   & $-26.18{\pm}2.18$  & $262.09{\pm}47.35$ & $0.18{\pm}.12$ & $0.10{\pm}.03$ & $0.00{\pm}.00$ & $0.06{\pm}.01$ & $0.00{\pm}.00$ & $0.16{\pm}.05$ & $2.0{\pm}2.6$ & $-0.30{\pm}.18$ \\
    FK-Steering  & $5.86{\pm}.30$ & $6.02{\pm}1.86$  & $-47.07{\pm}20.37$ & $128.04{\pm}16.51$ & $0.14{\pm}.06$ & $0.01{\pm}.01$ & $0.03{\pm}.04$ & $0.27{\pm}.03$ & $0.09{\pm}.01$ & $0.20{\pm}.03$ & $6.9{\pm}.7$ & $-0.18{\pm}.02$ \\
    TreeG        & $6.00{\pm}.34$ & $8.40{\pm}1.13$  & $-79.32{\pm}18.79$ & $110.77{\pm}20.73$ & $0.08{\pm}.03$ & $0.05{\pm}.01$ & $0.01{\pm}.01$ & $0.10{\pm}.01$ & $0.10{\pm}.01$ & $0.10{\pm}.01$ & $5.0{\pm}.4$ & $-0.24{\pm}.02$ \\
    DSearch      & $6.61{\pm}.33$ & $9.90{\pm}1.50$  & $-54.85{\pm}15.04$ & $120.11{\pm}27.84$ & $0.29{\pm}.02$ & $0.26{\pm}.03$ & $0.05{\pm}.05$ & $0.47{\pm}.03$ & $0.20{\pm}.02$ & $0.26{\pm}.01$ & $7.5{\pm}.8$ & $-0.21{\pm}.01$ \\
    Flow-Direct  & $\mathbf{7.18}{\pm}.27$ & $\mathbf{10.94}{\pm}1.06$ & $\mathbf{-14.63}{\pm}7.99$  & $\mathbf{284.92}{\pm}28.09$ & $\mathbf{0.41}{\pm}.02$ & $\mathbf{0.42}{\pm}.02$ & $\mathbf{0.32}{\pm}.05$ & $\mathbf{0.52}{\pm}.07$ & $\mathbf{0.25}{\pm}.04$ & $\mathbf{0.30}{\pm}.01$ & $\mathbf{9.0}{\pm}.5$ & $\mathbf{-0.16}{\pm}.01$ \\
    \bottomrule
  \end{tabular}
  }
\end{table}

\begin{table}[ht]
  \caption{Efficiency gain factor of Flow-Direct over each baseline. A factor of $k$ indicates that Flow-Direct matches or surpasses the baseline's final reward using $k$ times fewer reward evaluations.}
  \label{tab:result_eff}
  \centering
  \resizebox{\textwidth}{!}{%
  \begin{tabular}{lcccccccccccc}
    \toprule
    \multirow{2}{*}{Algorithm} & \multirow{2}{*}{Aesthetic} & \multirow{2}{*}{HPSv3} & \multirow{2}{*}{Compressibility} & \multirow{2}{*}{Incompressibility} & \multicolumn{6}{c}{Attribute Alignment} & \multirow{2}{*}{Prompt Alignment} & \multirow{2}{*}{Aerodynamic} \\
    \cmidrule(lr){6-11}
     & & & & & Happy & Fluffy & Running & Cute & Vivid & Fierce & & \\
    \midrule
    Evolvable   & $14.52\times$ & $45.40\times$  & $7.93\times$   & $3.07\times$   & $12.04\times$  & $21.43\times$   & $40.00\times$  & $83.33\times$  & $100.00\times$  & $25.00\times$  & $50.00\times$ & $77.78\times$ \\
    FK-Steering & $61.70\times$ & $100.00\times$ & $16.02\times$  & $25.00\times$  & $16.39\times$  & $100.00\times$  & $10.07\times$  & $27.78\times$  & $25.00\times$   & $23.33\times$  & $4.09\times$  & $3.14\times$  \\
    TreeG       & $49.48\times$ & $46.94\times$  & $54.17\times$  & $54.63\times$  & $26.11\times$  & $27.78\times$   & $23.61\times$  & $66.67\times$  & $20.56\times$   & $50.00\times$  & $13.06\times$ & $31.48\times$ \\
    DSearch     & $5.81\times$  & $5.13\times$   & $20.50\times$  & $39.24\times$  & $3.62\times$   & $5.86\times$    & $8.13\times$   & $2.03\times$   & $2.15\times$    & $5.76\times$   & $2.78\times$  & $6.64\times$  \\
    \bottomrule
  \end{tabular}%
  }
\end{table}

\begin{figure}[ht]
  \centering
  \includegraphics[width=0.8\linewidth]{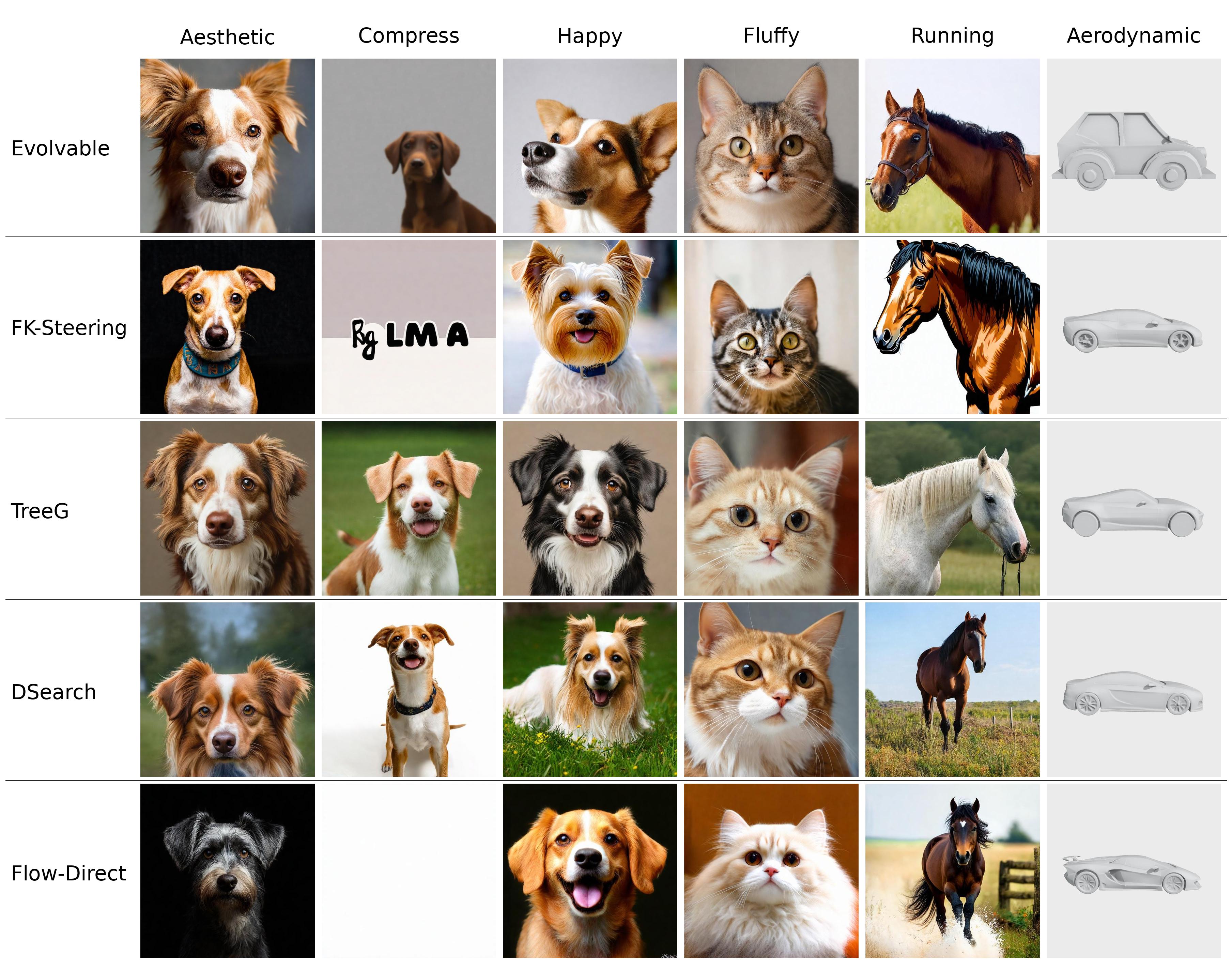}
  \caption{Samples generated by each method across six representative reward functions. Flow-Direct produces images that clearly reflect the target reward while preserving image quality.}
  \label{fig:result}
\end{figure}

\textbf{Quantitative Results.} \cref{tab:result} shows the rewards achieved by all methods: for each run we record the average reward over the final batch of generated samples, and for each task we report the mean and standard deviation across runs. Flow-Direct consistently achieves the highest reward across all seven reward functions. \cref{tab:result_eff} shows the efficiency gain factor, computed as the reward evaluation budget divided by the number of reward evaluations Flow-Direct needed to match or surpass each baseline. A factor of $k$ indicates that Flow-Direct reaches the same reward as the baseline using $k$ times fewer reward evaluations.

In~\cref{app:best_of_n}, we additionally compare Flow-Direct against Best-of-N, which independently generates $1600$ samples from the pre-trained model and selects the best sample. \cref{tab:best_of_n} shows Flow-Direct's final-batch mean already surpasses Best-of-N's maximum on 9 of 10 tasks, and significantly dominates on all 10 tasks under best-selection. This indicates that Flow-Direct optimizes the generative distribution, rather than merely retrieving the best sample from the accumulated dataset.

\textbf{Qualitative Results.} \cref{fig:result} visualizes samples generated by each method across six representative reward functions, spanning both 2D image alignment and 3D vehicle aerodynamic optimization. Additional images and 3D vehicle results are provided in \cref{app:additional_qualitative_image,app:additional_qualitative_3d}, respectively. Across both domains, Flow-Direct consistently produces samples that strongly reflect the target reward, while preserving the visual and structural quality. In contrast, baseline methods are less efficient at achieving these targets within the fixed reward evaluation budget.

\textbf{Ablation Study.}
Flow-Direct (Algorithm~\ref{alg:main}) has only two hyperparameters: the number of optimization iterations $L$ and the batch size $N$. We conduct an ablation study on these parameters in~\cref{app:ablation}. Results show that increasing either the optimization steps or the batch size yields a steady and stable improvement in the final reward, demonstrating the stability and scalability of Flow-Direct. Furthermore, unlike baseline methods, our formulation is compatible with both deterministic ODE and stochastic SDE samplers. In~\cref{tab:ablation_ode_sde}, we further compare Flow-Direct with the SDE sampler and the deterministic ODE sampler under the same reward-evaluation budget. The two variants achieve comparable rewards across reward functions, and both outperform all baselines, indicating that Flow-Direct is flexible and sampler-agnostic. Additionally, we analyze the computational overhead in \cref{app:ablation}. The results show that the computational time and memory consumption of the guidance field $\Delta(x_t)$ both scale linearly with the dataset size $N$.

\subsection{Reusable Target Datasets}

\begin{figure}[ht]
  \centering
  \includegraphics[width=0.8\linewidth]{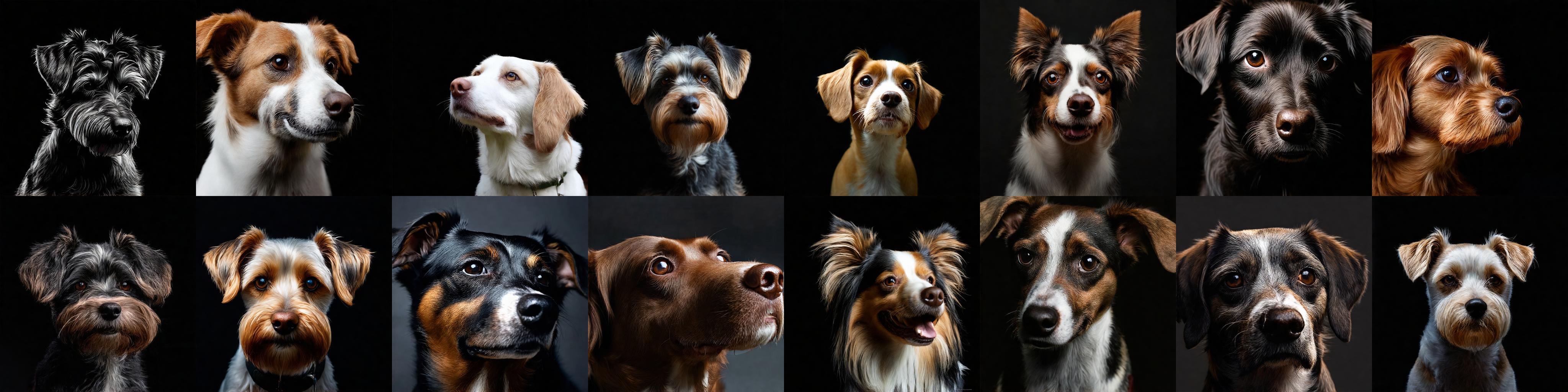}\\[2pt]
  \includegraphics[width=0.8\linewidth]{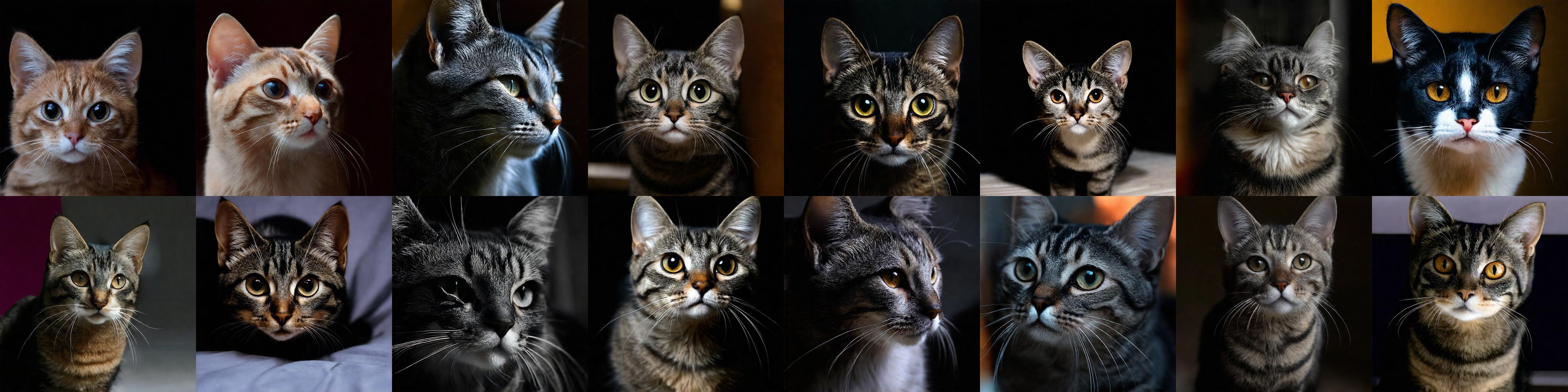}
\caption{Novel and diverse images generated \underline{without additional reward evaluation} by reusing the labeled dataset collected from the \textit{Aesthetic}-optimized \textit{dog} dataset. \textbf{Top:} prompt \textit{dog} (matching the optimization). \textbf{Bottom:} prompt \textit{cat} (unseen during optimization). The \textit{Aesthetic} style transfers in both cases, demonstrating the generalization capability.}
  \label{fig:reuse}
\end{figure}

We demonstrate that labeled datasets accumulated during optimization can be reused for target generation without additional reward evaluations. We construct a reusable guidance field by~\cref{eq:reuse_guidance} with $\alpha=1$ using the dataset collected during main experiment (\cref{tab:result}) with \textit{Aesthetic} task and \textit{dog} prompt. We then generate $1600$ independent samples with the original (\textit{dog}) and an unseen (\textit{cat}) prompt, respectively. Note that no additional reward evaluation is used during this reusable guided generation. \cref{tab:reuse} shows that the guidance field improves both the mean and the best reward over the pre-trained model, even on the unseen prompt, and \cref{fig:reuse} shows that the target style successfully transfers to both the original (\textit{dog}) and an unseen (\textit{cat}) prompt while preserving its visual quality.

\begin{table}[ht]
  \caption{Mean and best reward value over $1600$ independent samples generated by Flow-Direct's reusable guidance field. The reusable guidance field is constructed using the dataset collected during main experiment (\cref{tab:result}) with \textit{Aesthetic} task and \textit{dog} prompt. \underline{No additional reward evaluation} is used during the reusable run.}
  \label{tab:reuse}
  \centering
  \small
  \begin{tabular}{llcccc}
    \toprule
    \multirow{2}{*}{Reward} & \multirow{2}{*}{Method} & \multicolumn{2}{c}{\textit{Dog} (seen)} & \multicolumn{2}{c}{\textit{Cat} (unseen)} \\
    \cmidrule(lr){3-4} \cmidrule(lr){5-6}
     & & Mean & Best & Mean & Best \\
    \midrule
    \multirow{2}{*}{Aesthetic}       & Pre-trained            & $5.68$ & $6.99$ & $5.54$ & $6.71$ \\
                                     & Flow-Direct (reusable) & $\mathbf{6.44}$ & $\mathbf{7.41}$ & $\mathbf{6.13}$ & $\mathbf{7.10}$ \\
    \midrule
    \multirow{2}{*}{Compressibility} & Pre-trained            & $-86.12$ & $-41.91$ & $-84.02$ & $-43.04$ \\
                                     & Flow-Direct (reusable) & $\mathbf{-8.65}$ & $\mathbf{-6.68}$ & $\mathbf{-9.02}$ & $\mathbf{-6.42}$ \\
    \bottomrule
  \end{tabular}
\end{table}

\begin{figure}[ht]
  \centering
  \includegraphics[width=1\linewidth]{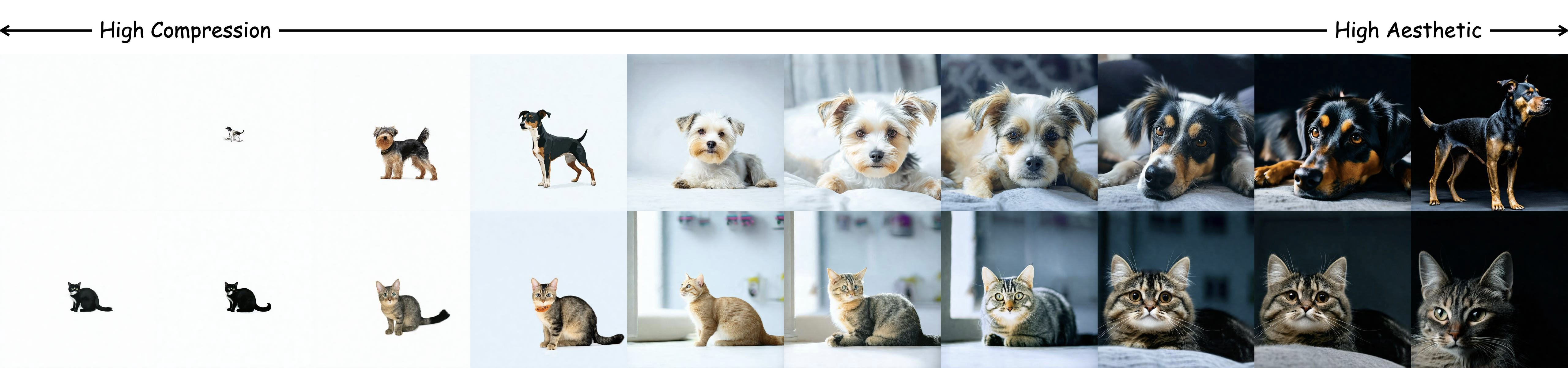}
  \caption{Compositional generation achieved \underline{without additional reward evaluation} by mixing datasets optimized for \textit{Aesthetic} and \textit{Compressibility} on the \textit{dog} prompt. \textbf{Top:} \textit{dog} prompt (matching the optimization). \textbf{Bottom:} \textit{cat} prompt (unseen during optimization). Both rows demonstrate a smooth interpolation between the two targets.}
  \label{fig:reuse_mix}
\end{figure}

Furthermore, multiple labeled datasets can be combined for multi-target generation. We apply \cref{eq:reuse_multi} using datasets optimized for the \textit{Aesthetic} and \textit{Compressibility} rewards on the \textit{dog} prompt. We interpolate their scalar weights along $\alpha_{\text{aes}} + \alpha_{\text{com}} = 1$ to balance the trade-off. As seen in \cref{fig:reuse_mix}, the images transition smoothly from high-compression to high-aesthetic. Additionally, this linear combination of distinct guidance fields can generalize to the unseen (\textit{cat}) prompt, indicating its generalizability.

\section{Conclusion, Limitations, and Broader Impacts}

\textbf{Conclusion.} We presented Flow-Direct, a training-free framework that addresses the fundamental feedback inefficiency of existing inference-time guidance methods. By analytically deriving a guidance field from the log-density ratio of the base and target distributions and implementing it via a non-parametric estimator, Flow-Direct persistently aggregates all reward-evaluated samples. This approach is highly feedback-efficient because black-box feedback is preserved rather than transiently discarded. Our experiments across image alignment and 3D aerodynamic optimization demonstrate Flow-Direct's superior feedback-efficiency across diverse domains.

\textbf{Limitations.} While Flow-Direct is highly efficient in terms of reward evaluations, it is an iterative algorithm that requires executing the full generation trajectory multiple times. Consequently, unlike most existing guidance algorithms that perform guidance during a single generation pass, our approach incurs a higher computational cost with respect to neural function evaluations (NFEs). Therefore, Flow-Direct is intended for the expensive-reward regime where query-efficiency is the bottleneck. Additionally, because the framework assumes a continuous state space and differentiable score function to formulate the guidance vector field, it is not applicable to discrete diffusion models.

\textbf{Broader Impacts.} Flow-Direct may improve the efficiency of reward-guided generation for scientific and design applications, but it also inherits broader risks of generative models, including misuse in realistic synthetic content generation.

\begin{ack}
This research is partly supported by the National Research Foundation, Singapore under its NRF AI-for-Science (AI4S) Challenge Grant (Award NRF-AI4SCH-2025-0007). Any opinions, findings and conclusions or recommendations expressed in this material are those of the author(s) and do not reflect the views of National Research Foundation, Singapore.
\end{ack}

\bibliographystyle{unsrtnat}
\bibliography{reference}

%%%%%%%%%%%%%%%%%%%%%%%%%%%%%%%%%%%%%%%%%%%%%%%%%%%%%%%%%%%%

\appendix

\newpage

\section{Derivations and Proofs}
\label[appendix]{app:proofs}

\subsection{Preliminaries: Flow Matching Model}
\label{app:preliminaries}

Given a noise distribution $x_0 \sim p_0 = \mathcal{N}(0, I)$ and a data distribution $x_1 \sim p_1$, flow matching framework~\citep{lipman2022flow} constructs a probability path between them via linear interpolation:
\begin{align}
x_t = (1-t)x_0 + tx_1.
\label{eq:app_interpolation}
\end{align}
Conditional on a data point $x_1$, the intermediate state $x_t$ is Gaussian:
\begin{align}
p_t(x_t | x_1) &= \mathcal{N}\big(x_t \big| tx_1, (1-t)^2 I\big) \\
&= \frac{1}{(2\pi(1-t)^2)^{D/2}} \exp\left(-\frac{\|x_t - tx_1\|^2}{2(1-t)^2}\right).
\label{eq:app_gaussian}
\end{align}
Marginalizing over $x_1$ yields the marginal density along the probability path,
\begin{align}
p_t(x_t) = \int p_t(x_t | x_1) p_1(x_1) dx_1,
\label{eq:app_marginal}
\end{align}
and applying Bayes' rule yields the posterior over the clean data conditioned on $x_t$:
\begin{align}
p_1(x_1 | x_t) = \frac{p_t(x_t | x_1) p_1(x_1)}{p_t(x_t)} \propto p_1(x_1) \exp\left(-\frac{\|x_t - tx_1\|^2}{2(1-t)^2}\right).
\label{eq:app_posterior}
\end{align}

\citet[Eq.~8]{lipman2022flow} shows that the marginal probability path $p_t(x_t)$ is generated by a \emph{marginal vector field} given as:
\begin{align}
v_t(x_t) 
&= \int v_t(x_t | x_1) \frac{p_t(x_t | x_1) p_1(x_1)}{p_t(x_t)}\, dx_1 \\
&= \int \frac{x_1 - x_t}{1-t} p_1(x_1 | x_t) dx_1 \\
&= \frac{1}{1-t}\left( \int x_1 p_1(x_1 | x_t) dx_1 - x_t \int p_1(x_1 | x_t) dx_1 \right) \\
&= \frac{\mathbb{E}[x_1 | x_t] - x_t}{1-t},
\label{eq:app_vt}
\end{align}
where $v_t(x_t | x_1)$ is the conditional vector field that generates $p_t(x_t | x_1)$. The second line applies Bayes' rule from \cref{eq:app_posterior}, and the last line uses $\int p_1(x_1 \mid x_t)\, dx_1 = 1$.

\subsection{Score--Vector Field Identity (Derivation of~\cref{eq:vector_score})}
\label{app:score_identity}

We now derive the identity in~\cref{eq:vector_score}:
\begin{align}
v_t(x_t) = \frac{x_t}{t} + \frac{1-t}{t} \nabla_{x_t} \log p_t(x_t),
\end{align}
which expresses the marginal vector field $v_t$ in terms of the score of the probability path $p_t(x_t)$.

\paragraph{Score Function as Posterior Expectation.}
We first express the score function in terms of the posterior expectation by taking gradient of the marginal log-density defined in \cref{eq:app_marginal}:
\begin{align}
\nabla_{x_t} \log p_t(x_t)
&= \frac{1}{p_t(x_t)} \nabla_{x_t} \int p_t(x_t | x_1) p_1(x_1) dx_1 \notag \\
&= \frac{1}{p_t(x_t)} \int p_t(x_t | x_1) \nabla_{x_t} \log p_t(x_t | x_1) p_1(x_1) dx_1 \notag \\
&= \int p_1(x_1 | x_t) \nabla_{x_t} \log p_t(x_t | x_1) dx_1 \notag \\
&= \int p_1(x_1 | x_t) \frac{tx_1 - x_t}{(1-t)^2} dx_1 \notag \\
&= \frac{t \mathbb{E}[x_1 | x_t] - x_t}{(1-t)^2}, \notag \\
\implies \mathbb{E}[x_1 | x_t] &= \frac{x_t}{t} + \frac{(1-t)^2}{t} \nabla_{x_t} \log p_t(x_t),
\label{eq:app_posterior_mean}
\end{align}
where the first line applies the log-derivative identity $\nabla \log p = \nabla p / p$ and substitutes the marginal from \cref{eq:app_marginal}, the second line moves $\nabla_{x_t}$ inside the integral and reapplies $\nabla p = p \nabla \log p$ to the conditional, the third line applies Bayes' rule from \cref{eq:app_posterior}, the fourth line differentiates the log-density of Gaussian in~\cref{eq:app_gaussian} with respect to $x_t$, and the fifth line uses the linearity of expectation. The last line rearranges in terms of the posterior expectation.

\paragraph{Vector field in terms of the score.} Substituting \cref{eq:app_posterior_mean} into the marginal vector field in \cref{eq:app_vt},
\begin{align}
v_t(x_t)
&= \frac{\mathbb{E}[x_1 | x_t] - x_t}{1-t} \notag \\
&= \frac{1}{1-t}\left( \frac{x_t}{t} + \frac{(1-t)^2}{t} \nabla_{x_t} \log p_t(x_t) - x_t \right) \notag \\
&= \frac{x_t}{t} + \frac{1-t}{t} \nabla_{x_t} \log p_t(x_t),
\end{align}
which recovers~\cref{eq:vector_score}. Specializing this identity to the probability path $p_t^\theta$ generated by the pre-trained model and assuming the model is perfectly well-trained $v_\theta(x_t) \approx v_t(x_t)$ yields the~\cref{eq:vector_score_theta}, completing the derivation.

\subsection{Guided Vector Field as a Posterior-Expectation Difference (Derivation of~\cref{eq:guided_flow_2})}
\label{app:shift_derivation}

In~\cref{eq:guided_flow_1}, the gradient of the log-density ratio between the two probability paths can be written as:
\begin{align}
\nabla_{x_t} \log \frac{p^\text{target}_t(x_t)}{p^\text{base}_t(x_t)}
&= \nabla_{x_t} \log p^\text{target}_t(x_t) - \nabla_{x_t} \log p^\text{base}_t(x_t) \notag \\
&= \frac{t\, \mathbb{E}_{p_1^\text{target}}[x_1 | x_t] - x_t}{(1-t)^2} - \frac{t\, \mathbb{E}_{p_1^\text{base}}[x_1 | x_t] - x_t}{(1-t)^2} \notag \\
&= \frac{t}{(1-t)^2} \left( \mathbb{E}_{p_1^\text{target}}[x_1 | x_t] - \mathbb{E}_{p_1^\text{base}}[x_1 | x_t] \right),
\label{eq:app_shift_score}
\end{align}
where the second line applies the marginal-score identity from \cref{eq:app_posterior_mean} to each probability path, and the $x_t$ terms cancel in the difference. Substituting \cref{eq:app_shift_score} into \cref{eq:guided_flow_1} yields:
\begin{align}
v^\text{guided}_\theta(x_t) 
&= \frac{x_t}{t} + \frac{1-t}{t} \left( \nabla_{x_t} \log p_t^\theta(x_t) + \frac{t}{(1-t)^2} \left( \mathbb{E}_{p_1^\text{target}}[x_1 | x_t] - \mathbb{E}_{p_1^\text{base}}[x_1 | x_t] \right) \right) \notag \\
&= v_\theta(x_t) + \frac{1}{1-t} \left( \mathbb{E}_{p_1^\text{target}}[x_1 | x_t] - \mathbb{E}_{p_1^\text{base}}[x_1 | x_t] \right),
\end{align}
which recovers~\cref{eq:guided_flow_2} and completes the derivation.

\subsection{Proof of Proposition 1 (Estimator for Posterior Expectation)}

Let $\mathcal{D} = \{x_1^i\}_{i=1}^N$ be a finite dataset of $N$ samples drawn from the true data distribution $p_1(x_1)$, i.e., $x_1^i \overset{\text{iid}}{\sim} p_1(x_1)$. We approximate this unknown distribution using the empirical distribution $\hat{p}_1(x_1)$, represented as a uniform mixture of Dirac delta functions:
\begin{align}
\hat{p}_1(x_1) \coloneqq \frac{1}{N} \sum_{i=1}^N \delta(x_1 - x_1^i) \xrightarrow{N \to \infty} p_1(x_1).
\label{eq:empirical_distribution}
\end{align}
By the Glivenko-Cantelli theorem, the empirical measure $\hat{p}_1(x_1)$ converges to the true underlying data distribution $p_1(x_1)$ as the dataset size approaches infinity.

Following~\citep{bertrand2025closed,kamb2024analytic}, under the empirical data distribution, the empirical posterior expectation is:
\begin{align}
\mathbb{E}_{\hat{p}_1}[x_1 | x_t] &= \frac{\int p_t(x_t | x_1) \hat{p}_1(x_1) x_1 dx_1}{\int p_t(x_t | x_1) \hat{p}_1(x_1) dx_1} \notag \\
&= \frac{ \int \mathcal{N}(x_t | t x_1, (1-t)^2 I) \big[ \frac{1}{N} \sum_{i=1}^N \delta(x_1 - x_1^i) \big] x_1 dx_1 }{ \int \mathcal{N}(x_t | t x_1, (1-t)^2 I) \big[ \frac{1}{N} \sum_{i'=1}^N \delta(x_1 - x_1^{i'}) \big] dx_1 } \notag \\
&= \frac{ \sum_{i=1}^N \mathcal{N}(x_t | t x_1^i, (1-t)^2 I) x_1^i }{ \sum_{i'=1}^N \mathcal{N}(x_t | t x_1^{i'}, (1-t)^2 I) } \notag \\
&= \frac{ \sum_{i=1}^N \exp \left( -\frac{\|x_t - t x_1^i\|^2}{2(1-t)^2} \right) x_1^i }{ \sum_{i'=1}^N \exp \left( -\frac{\|x_t - t x_1^{i'}\|^2}{2(1-t)^2} \right) } \notag \\
&= \sum_{i=1}^N \softmax_{i' \in [N]}\left(-\frac{\|x_t - t x_1^{i'}\|^2}{2(1-t)^2}\right)_{\!\!i} x_1^i.
\label{eq:empirical_posterior_expectation}
\end{align}

The first line expands the posterior expectation using Bayes' rule. In the second line, we substitute the conditional Gaussian density $p_t(x_t | x_1)$ and the empirical data distribution $\hat{p}_1(x_1)$. In the third line, by applying the sifting property of the Dirac delta function ($\int f(x)\delta(x-x_0)dx = f(x_0)$), the continuous integrals evaluate exactly to discrete sums over the dataset $\mathcal{D}$. The fourth line expands the Gaussian probability densities, which allows the shared normalization constant to cancel out and form the Softmax. When dataset size approaches infinity, then $\mathbb{E}_{\hat{p}_1}[x_1 | x_t] \rightarrow \mathbb{E}_{p_1}[x_1 | x_t]$, which recovers~\cref{eq:posterior_estimator} and completes the proof.

\subsection{Proof of Proposition 2 (Estimator for Target Posterior Expectation)}

Recall the target distribution is defined as:
\begin{align}
p^\text{target}_1(x_1) \propto p^\text{base}_1(x_1)\exp(r(x_1)).
\label{eq:app_target_distribution}
\end{align}
Similarly, let $\mathcal{D} = \{x_1^i\}_{i=1}^{N}$ be a finite dataset of $N$ samples drawn from the base distribution $p_1^\text{base}(x_1)$, i.e., $x_1^i \overset{\text{iid}}{\sim} p_1^\text{base}(x_1)$. We approximate the target distribution using the empirical target distribution:
\begin{align}
\hat{p}_1^\text{target}(x_1) \coloneqq \frac{1}{\hat{Z} N} \sum_{i=1}^N \exp(r(x_1^i)) \delta(x_1 - x_1^i) \xrightarrow{N \to \infty} p_1^\text{target}(x_1),
\end{align}
where $\hat{Z}$ is the empirical normalizing constant, and $r(x_1)$ is reward function. By the consistency of self-normalized importance sampling, the empirical measure $\hat{p}^\text{target}_1(x_1)$ converges to the true target data distribution $p^\text{target}_1(x_1)$ as the dataset size approaches infinity.

Following the similar derivation as~\cref{eq:empirical_posterior_expectation}, the posterior expectation under this empirical target distribution is:
\begin{align}
\mathbb{E}_{\hat{p}_1^\text{target}}[x_1 | x_t] &= \frac{\int  p_t(x_t | x_1) \hat{p}_1^\text{target}(x_1) x_1 dx_1}{\int p_t(x_t | x_1) \hat{p}_1^\text{target}(x_1) dx_1} \notag \\
&= \frac{ \int  \mathcal{N}(x_t | t x_1, (1-t)^2 I) \big[ \frac{1}{\hat{Z}N} \sum_{i=1}^N \exp(r(x_1^i)) \delta(x_1 - x_1^i) \big] x_1 dx_1 }{ \int \mathcal{N}(x_t | t x_1, (1-t)^2 I) \big[ \frac{1}{\hat{Z}N} \sum_{i'=1}^N \exp(r(x_1^{i'})) \delta(x_1 - x_1^{i'}) \big] dx_1 } \notag \\
&= \frac{ \sum_{i=1}^N \mathcal{N}(x_t | t x_1^i, (1-t)^2 I) \exp(r(x_1^i)) x_1^i }{ \sum_{i'=1}^N \mathcal{N}(x_t | t x_1^{i'}, (1-t)^2 I) \exp(r(x_1^{i'})) } \notag \\
&= \frac{ \sum_{i=1}^N \exp \left( -\frac{\|x_t - t x_1^i\|^2}{2(1-t)^2} + r(x_1^i) \right) x_1^i }{ \sum_{i'=1}^N \exp \left( -\frac{\|x_t - t x_1^{i'}\|^2}{2(1-t)^2} + r(x_1^{i'}) \right) } \notag \\
&= \sum_{i=1}^N \softmax_{i' \in [N]}\left(-\frac{\|x_t - t x_1^{i'}\|^2}{2(1-t)^2} + r(x_1^{i'}) \right)_{\!\!i} x_1^i.
\end{align}
which recovers~\cref{eq:target_posterior_estimator} and completes the proof.

\section{Practical Implementation}
\label[appendix]{app:implementation}

We now discuss several practical considerations for implementing the posterior expectations in~\cref{proposition:1,proposition:2}. Since both estimators share the same functional form, and the latter is simply the reward-weighted extension of the former, we focus our discussion on the latter; the following techniques apply equally to both. Recall the target posterior estimator in~\cref{eq:target_posterior_estimator}:
\begin{align}
\mathbb{E}_{p^\textnormal{target}_1}[x_1 | x_t] \approx 
\sum_{i=1}^N
\softmax_{i' \in [N]}\!\left(-\frac{\|x_t-tx^{i'}_1\|^2}{2(1-t)^2} + r_{i'}\right)_{\!\!i} x_1^i.
\tag{\ref{eq:target_posterior_estimator}}
\end{align}

The Softmax logits dictate the influence of each data point by combining its $\ell_2$ distance from the current state with its scalar reward. To prevent numerical instability during optimization, the distance metric must remain robust across high dimensionalities and shifting timesteps. Furthermore, the distance and reward terms must be calibrated to a similar scale so that neither dominates the guidance signal. We address these requirements through the following practical techniques:

\begin{itemize}[leftmargin=2.5em]
\item In high-dimensional space $\mathbb{R}^D$, the standard deviation of the squared $\ell_2$ distance between two independent vectors scales with $\sqrt{D}$. We therefore normalize the squared distance term by $\sqrt{D}$ to ensure the variance for the Softmax logits is dimension-invariant.

\item
Although $tx^{i'}_1 = \mathbb{E}[x_t^{i'}|x_1^{i'}]$ is the correct conditional mean, its norm shrinks toward the origin for $t<1$, placing it on a different scale than the noisy state $x_t$. For numerical stability, we replace the conditional mean with a single stochastic sample:
\begin{align}
	tx^{i'}_1 \leftarrow x_t^{i'} \coloneqq tx_1^{i'} + (1-t)\epsilon,
	\label{eq:noisy_expectation}
\end{align}
for $\epsilon \sim \mathcal{N}(0,I)$. This ensures that $x_t$ and $x_t^{i'}$ reside on the same scale.

\item
Under flow matching framework, the true posterior expectation must converge to current state, i.e., $\mathbb{E}[x_1 | x_t] \rightarrow x_t$, when $t \rightarrow 1$. To strictly enforce this property in our estimator, we append the model's prediction of the clean data to the dataset:
\begin{align}
	\tilde{\mathcal{D}} \leftarrow \mathcal{D} \cup \{\hat{x}_{1|t}\},
	\label{eq:regularized_dataset}
\end{align}
where $\hat{x}_{1|t} = x_t + (1-t) v_\theta(x_t)$ is the predicted clean data.

\item 
Building upon the previous technique, for the predicted clean data $\hat{x}_{1|t}$, we estimate its reward value using the existing dataset:
\begin{align}
\mathbb{E}_{p_1^\text{base}}[r(x_1)|x_t] \approx \sum_{i=1}^N
\softmax_{i' \in [N]} \left(-\frac{\|x_t - x_t^{i'}\|^2}{2(1-t)^2 \sqrt{D}}\right)_{\!\!i} r_i.
\label{eq:reward_estimator}
\end{align}
This reuses existing reward labels and incurs no additional reward evaluations.

\item Finally, we standardize the reward values to have zero mean and unit variance. For the iterative optimization described in~\cref{sec:3.2}, we normalize the rewards across all accumulated datasets ($\mathcal{D}_0, \mathcal{D}_1, \dots, \mathcal{D}_{l-1}$) using the same global mean and standard deviation. This ensures that the reward scale remains consistent across all telescoping guidance terms.

\end{itemize}

We study the effect of these techniques in the ablation study in~\cref{app:ablation_practical}.

Incorporating the above implementation techniques, we summarize the practical guidance fields used throughout the paper:\footnote{With a slight overload of notation, $N$ and $M$ now denote the sizes of the augmented datasets $\tilde{\mathcal{D}}$.}

\begin{itemize}[leftmargin=2.5em]
\item \textbf{Target Data Guidance (\cref{sec:3.1}):}
\begin{equation}
\label{eq:data_field_practical}
\resizebox{\linewidth}{!}{$\displaystyle
\boxed{
\Delta(x_t;\tilde{\mathcal{D}}_\text{base},\tilde{\mathcal{D}}_\text{target}) \coloneqq \frac{1}{1-t} \Bigg[
 \sum_{j=1}^{M} \softmax_{j' \in [M]} \left(
 -\frac{\|x_t - x_t^{j'}\|^2}{2(1-t)^2 \sqrt{D}}
 \right)_{\!\!j} x_1^{j}
 -
 \sum_{i=1}^{N} \softmax_{i' \in [N]} \left(
 -\frac{\|x_t - x_t^{i'}\|^2}{2(1-t)^2 \sqrt{D}}
 \right)_{\!\!i} x_1^{i}
\Bigg]
}
$}
\tag{\ref{eq:data_field}-practical}
\end{equation}
    
    \item \textbf{Iterative Reward Optimization (\cref{sec:3.2}):}
    \begin{equation}
    \label{eq:reward_guidance_practical}
    \resizebox{\linewidth}{!}{$\displaystyle
    \boxed{
    \Delta(x_t;\tilde{\mathcal{D}}) \coloneqq \frac{1}{1-t}
    \sum_{i=1}^{N} \left[
    \softmax_{i' \in [N]} \left(
    -\frac{\|x_t - x_t^{i'}\|^2}{2(1-t)^2 \sqrt{D}} + r_{i'}
    \right)_{\!\!i}
    -
    \softmax_{i' \in [N]} \left(
    -\frac{\|x_t - x_t^{i'}\|^2}{2(1-t)^2 \sqrt{D}}
    \right)_{\!\!i}
    \right] x_1^i
    }
    $}
    \tag{\ref{eq:reward_guidance}-practical}
    \end{equation}
    
    \item \textbf{Reusable Guidance Fields (\cref{sec:3.3}):}
    \begin{equation}
    \label{eq:reuse_guidance_practical}
    \resizebox{\linewidth}{!}{$\displaystyle
    \boxed{
    \Delta(x_t,\tilde{\mathcal{D}}_\text{all}, \alpha) \coloneqq \frac{1}{1-t}
    \sum_{i=1}^{N} \left[
    \softmax_{i' \in [N]} \left(
    -\frac{\|x_t - x_t^{i'}\|^2}{2(1-t)^2 \sqrt{D}} + \alpha r_{i'}
    \right)_{\!\!i}
    -
    \softmax_{i' \in [N]} \left(
    -\frac{\|x_t - x_t^{i'}\|^2}{2(1-t)^2 \sqrt{D}} - \alpha r_{i'}
    \right)_{\!\!i}
    \right] x_1^i
    }
    $}
    \tag{\ref{eq:reuse_guidance}-practical}
    \end{equation}
\end{itemize}

\section{Ablation Study}
\label[appendix]{app:ablation}

\subsection{Practical Implementation Techniques}
\label[appendix]{app:ablation_practical}

The practical implementation in~\cref{app:implementation} comprises four techniques:
\begin{enumerate}[leftmargin=2.5em]
\item Normalizing the softmax logits by $\sqrt{D}$.
\item Standardizing reward values to zero mean and unit variance.
\item Regularizing the dataset with the model's predicted clean sample (\cref{eq:regularized_dataset,eq:reward_estimator}).
\item Estimating the conditional expectation with a noisy sample (\cref{eq:noisy_expectation}).
\end{enumerate}

\textbf{Technique 1 is essential.} It is standard rather than novel to our method, following the logit scaling used in attention layers~\citep{vaswani2017attention} and RBF-kernel normalization~\citep{hvarfner2024vanilla}. Without it, the squared $\ell_2$ distance grows with the dimension $D$, so the softmax saturates to a near one-hot weighting over the dataset and the estimated posterior expectation collapses onto a single sample.

\textbf{Technique 2 is essential.} It is standard rather than novel to our method, and is common practice in reward-based generative optimization~\citep{liu2025flow,black2023training,xue2025advantage}. Raw reward values differ by orders of magnitude across tasks (e.g., Aesthetic $\in [1,10]$, Incompressibility $\approx 10^2$, Attribute Alignment $\in [0,1]$). Without it, the reward weighting could either over-concentrate on a few samples or become uniform, requiring a per-task scaling coefficient to be tuned.

\textbf{Technique 3 preserves perceptual image quality.} We observe that adding the model's predicted clean sample to the dataset can improve the perceptual quality of the generated images. This is reflected in~\cref{tab:ablation_practical}: with the regularized dataset, Flow-Direct performs better on the quality-oriented rewards (Aesthetic, HPSv3), but worse on the semantic rewards (Attribute Alignment). This is because appending the model's own predicted clean sample anchors the guidance field to the pre-trained data, which regularizes the guidance to remain near the pre-trained distribution.

\begin{table}[ht]
  \caption{Ablation study on practical implementation techniques (regularized dataset and noisy expectation estimation). The last row is the default Flow-Direct reported in~\cref{tab:result}. Note: As \textit{Compressibility} is a minimization task, its values are inverted.}
  \label{tab:ablation_practical}
  \centering
  \resizebox{\textwidth}{!}{%
  \begin{tabular}{cccccccccccc}
    \toprule
    Regularized & Noisy Expectation & \multirow{2}{*}{Aesthetic} & \multirow{2}{*}{HPSv3} & \multirow{2}{*}{Compressibility} & \multirow{2}{*}{Incompressibility} & \multicolumn{6}{c}{Attribute Alignment} \\
    \cmidrule(lr){7-12}
    Dataset (Technique 3) & Estimation (Technique 4) & & & & & Happy & Fluffy & Running & Cute & Vivid & Fierce \\
    \midrule
    \ding{55} & \ding{55} & $6.980{\pm}.131$ & $9.475{\pm}1.230$  & $-14.550{\pm}5.591$ & $290.547{\pm}11.676$ & $0.501{\pm}.064$ & $0.450{\pm}.014$ & $0.334{\pm}.090$ & $0.618{\pm}.066$ & $0.309{\pm}.010$ & $0.355{\pm}.027$ \\
    \ding{55} & \ding{51} & $6.968{\pm}.105$ & $9.425{\pm}1.075$  & $-12.897{\pm}6.068$ & $287.992{\pm}9.231$  & $0.478{\pm}.042$ & $0.432{\pm}.019$ & $0.273{\pm}.001$ & $0.700{\pm}.036$ & $0.312{\pm}.020$ & $0.352{\pm}.014$ \\
    \ding{51} & \ding{55} & $7.185{\pm}.270$ & $10.918{\pm}1.108$ & $-12.117{\pm}6.328$ & $295.753{\pm}30.628$ & $0.323{\pm}.018$ & $0.429{\pm}.021$ & $0.299{\pm}.050$ & $0.514{\pm}.041$ & $0.256{\pm}.019$ & $0.315{\pm}.015$ \\
    \ding{51} & \ding{51} & $7.181{\pm}.274$ & $10.944{\pm}1.055$ & $-14.632{\pm}7.993$ & $284.920{\pm}28.089$ & $0.406{\pm}.022$ & $0.424{\pm}.018$ & $0.324{\pm}.054$ & $0.521{\pm}.065$ & $0.253{\pm}.036$ & $0.300{\pm}.007$ \\
    \bottomrule
  \end{tabular}%
  }
\end{table}

\textbf{Technique 4 preserves sample diversity.} We visually observe that noisy estimation of the conditional expectation produces more diverse samples within the generated batch. To quantify this, we report the mean pairwise $\ell_2$ distance within each generated batch in~\cref{tab:ablation_diversity}. With noisy expectation estimation, the generated batch consistently yields a higher pairwise distance, indicating higher diversity, across all ten reward functions. We did not visually observe that noisy expectation estimation affects the image quality.

\begin{table}[ht]
  \caption{Pairwise $\ell_2$ distance on noisy expectation estimation.}
  \label{tab:ablation_diversity}
  \centering
  \resizebox{\textwidth}{!}{%
  \begin{tabular}{ccccccccccc}
    \toprule
    Noisy Expectation & \multirow{2}{*}{Aesthetic} & \multirow{2}{*}{HPSv3} & \multirow{2}{*}{Compressibility} & \multirow{2}{*}{Incompressibility} & \multicolumn{6}{c}{Attribute Alignment} \\
    \cmidrule(lr){6-11}
    Estimation (Technique 4) & & & & & Happy & Fluffy & Running & Cute & Vivid & Fierce \\
    \midrule
    \ding{55} & $93.9{\pm}7.8$ & $90.8{\pm}8.2$ & $34.3{\pm}32.0$ & $116.1{\pm}28.6$ & $91.4{\pm}4.2$ & $103.7{\pm}3.4$ & $110.2{\pm}6.9$ & $95.3{\pm}15.4$ & $86.3{\pm}2.1$ & $105.5{\pm}2.8$ \\
    \ding{51} & $95.8{\pm}9.2$ & $95.3{\pm}7.1$ & $50.6{\pm}48.1$ & $120.7{\pm}25.1$ & $98.8{\pm}2.5$ & $107.8{\pm}3.9$ & $115.5{\pm}9.9$ & $112.1{\pm}8.1$ & $91.4{\pm}3.1$ & $117.2{\pm}1.4$ \\
    \bottomrule
  \end{tabular}%
  }
\end{table}

\subsection{Optimization Steps and Batch Size}

We study the effect of the two hyperparameters in Flow-Direct: the number of optimization iterations $L$ and the batch size $N$. For each setting, we run six independent runs with distinct prompts to optimize the Aesthetic reward, and report the average reward.

\begin{figure}[htbp]
  \centering

  \begin{subfigure}{0.49\linewidth}
    \centering
    \includegraphics[width=\linewidth]{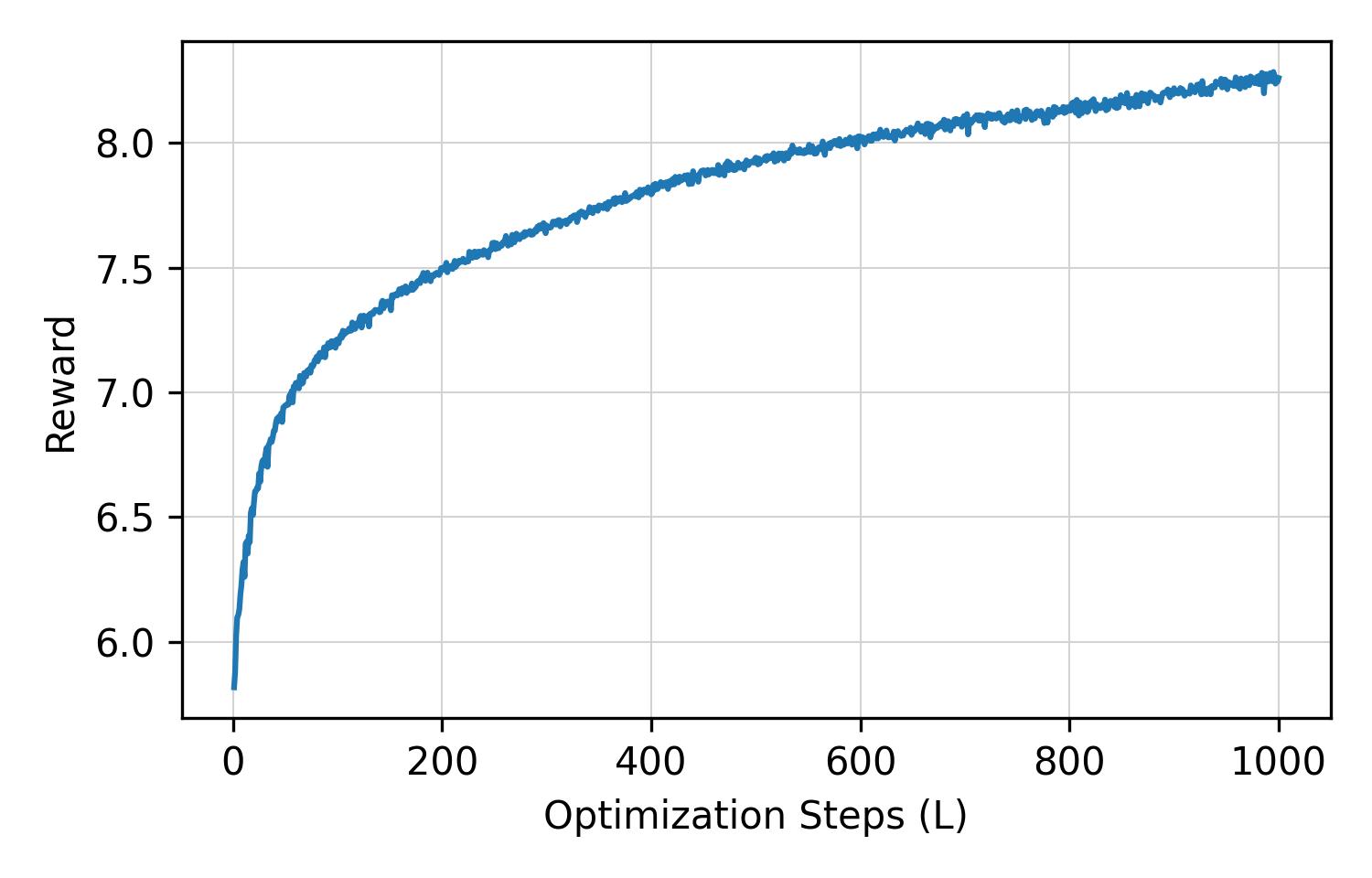}
    \caption{}
    \label{fig:ablation_steps}
  \end{subfigure}
  \begin{subfigure}{0.49\linewidth}
    \centering
    \includegraphics[width=\linewidth]{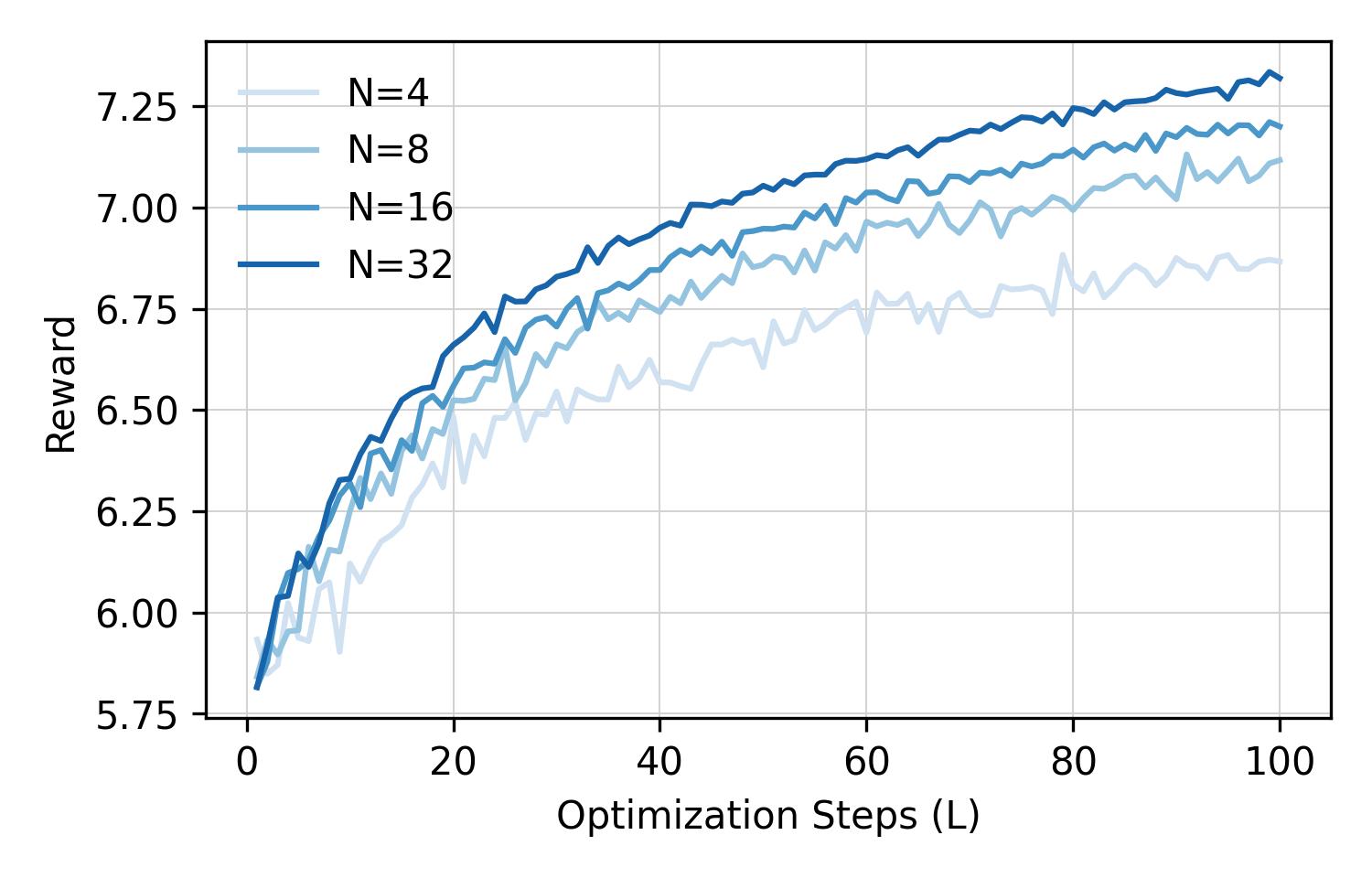}
    \caption{}
    \label{fig:ablation_batch}
  \end{subfigure}

  \caption{\textbf{Ablation study on scalability.}
  We report the average Aesthetic reward across six runs with distinct prompts.
  \textbf{(a)} Increasing the number of optimization steps continues to improve reward.
  \textbf{(b)} Larger batch sizes yield higher rewards.}
  \label{fig:ablation}
\end{figure}

In \cref{fig:ablation_steps}, we extend the optimization iteration to $L=1000$, which corresponds to $16{,}000$ reward evaluations when using batch size $N=16$. The reward increases steadily throughout optimization, indicating that Flow-Direct continues to benefit from additional reward-evaluated samples. In \cref{fig:ablation_batch}, we vary the batch size over $N \in \{4,8,16,32\}$. Larger batch sizes achieve higher rewards and produce more stable improvement curves. This is because each iteration contributes more labeled samples to the empirical dataset, yielding a more accurate posterior-expectation estimation. These results indicate both scalability and stability: Flow-Direct improves steadily as either the optimization horizon or batch size increases.

\subsection{ODE vs. SDE Sampler}
We study the effect of the sampler used by Flow-Direct. Specifically, we compare the performance of Flow-Direct under the SDE sampler with $\eta=0.7$ and the ODE sampler with $\eta=0.0$ in~\cref{eq:sde_step}. All other hyperparameters and the reward-evaluation budget are kept the same as in the main experiment. We report the results in~\cref{tab:ablation_ode_sde}. We observe that both variants achieve comparable rewards across all reward functions. Moreover, both Flow-Direct (SDE) and Flow-Direct (ODE) achieve higher rewards than all baseline methods reported in~\cref{tab:result}, showing that Flow-Direct is effective under both stochastic and deterministic samplers.

\begin{table}[ht]
\caption{Reward value achieved by Flow-Direct with SDE and ODE samplers.}
\label{tab:ablation_ode_sde}
\centering
\resizebox{\textwidth}{!}{%
\begin{tabular}{lccccccccccc}
\toprule
\multirow{2}{*}{Algorithm} & \multirow{2}{*}{Aesthetic} & \multirow{2}{*}{HPSv3} & \multirow{2}{*}{Compressibility} & \multirow{2}{*}{Incompressibility} & \multicolumn{6}{c}{Attribute Alignment} & \multirow{2}{*}{Aerodynamic} \\
\cmidrule(lr){6-11}
& & & & & Happy & Fluffy & Running & Cute & Vivid & Fierce & \\
\midrule
Flow-Direct (SDE) & $\mathbf{7.18}{\pm}0.27$ & $10.94{\pm}1.05$ & $-14.63{\pm}7.99$ & $\mathbf{284.92}{\pm}28.09$ & $\mathbf{0.41}{\pm}0.02$ & $0.42{\pm}0.02$ & $\mathbf{0.32}{\pm}0.05$ & $0.52{\pm}0.07$ & $0.25{\pm}0.04$ & $\mathbf{0.30}{\pm}0.01$ & $-0.16{\pm}0.01$ \\
Flow-Direct (ODE) & $7.15{\pm}0.29$ & $\mathbf{11.26}{\pm}1.18$ & $\mathbf{-8.77}{\pm}6.66$ & $273.07{\pm}33.69$ & $\mathbf{0.41}{\pm}0.03$ & $\mathbf{0.45}{\pm}0.01$ & $0.23{\pm}0.10$ & $\mathbf{0.58}{\pm}0.07$ & $\mathbf{0.27}{\pm}0.02$ & $0.29{\pm}0.00$ & $\mathbf{-0.14}{\pm}0.01$ \\
\bottomrule
\end{tabular}%
}
\end{table}

\subsection{Computational Resource}

\begin{figure}[ht]
  \centering

  \begin{subfigure}{0.32\linewidth}
    \centering
    \includegraphics[width=\linewidth]{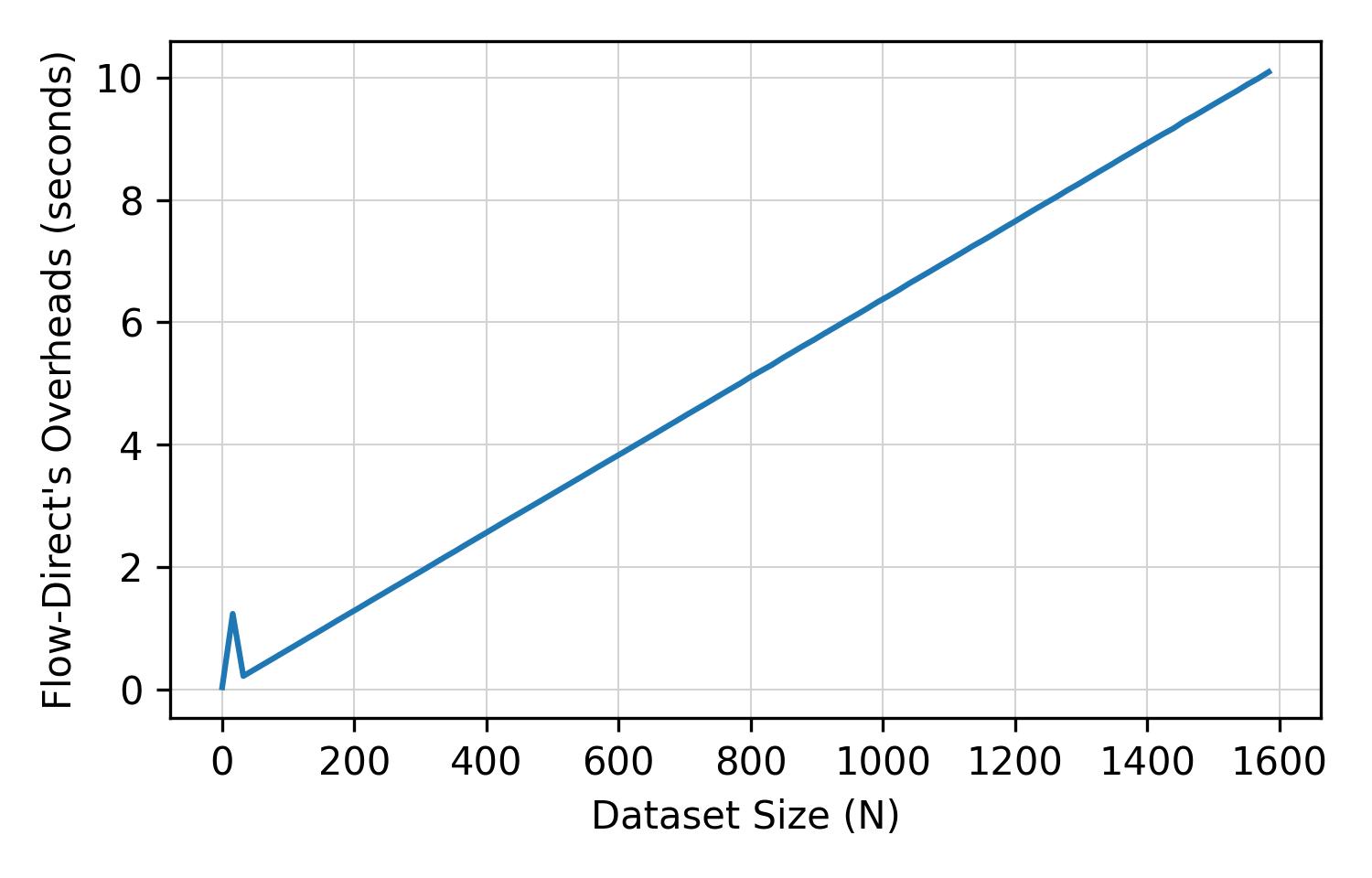}
    \caption{}
    \label{fig:ablation_time_seconds}
  \end{subfigure}
  \begin{subfigure}{0.32\linewidth}
    \centering
    \includegraphics[width=\linewidth]{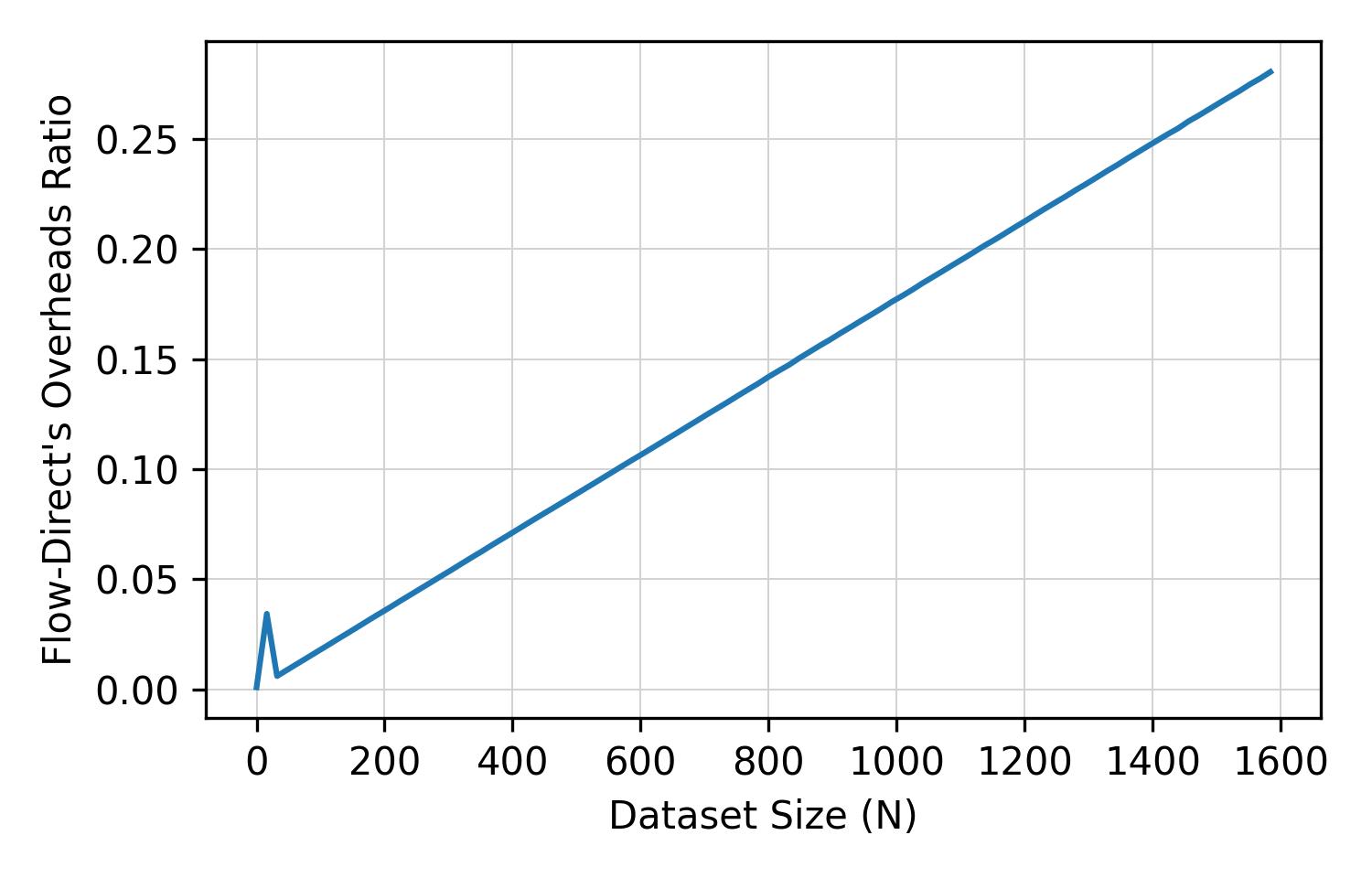}
    \caption{}
    \label{fig:ablation_time_ratio}
  \end{subfigure}
  \begin{subfigure}{0.32\linewidth}
    \centering
    \includegraphics[width=\linewidth]{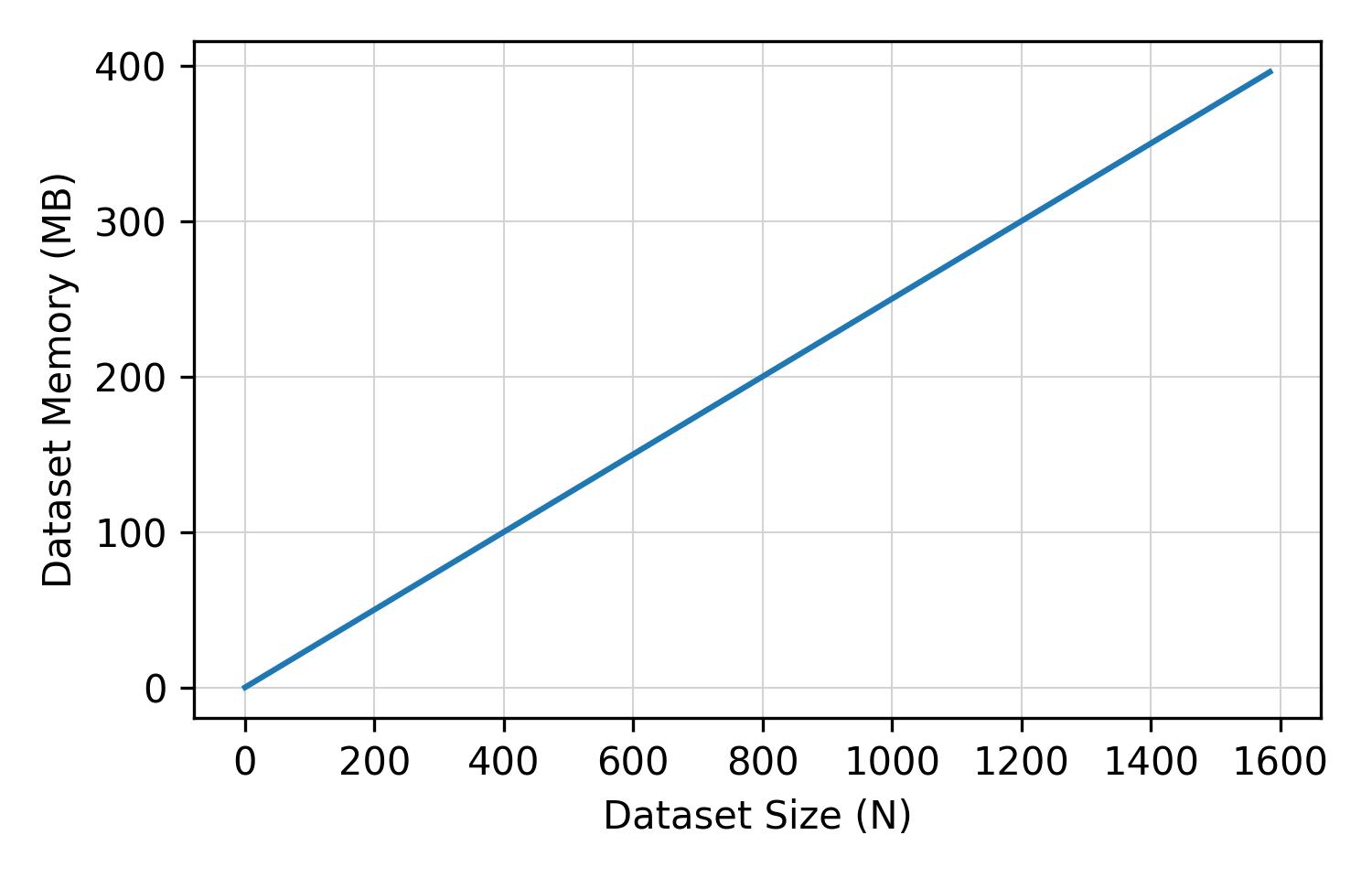}
    \caption{}
    \label{fig:ablation_time_memory}
  \end{subfigure}
  \caption{\textbf{Ablation study on computational overhead and memory.} Flow-Direct's computation time and memory consumption scale linearly with the dataset size $N$.}
  \label{fig:ablation_computation}
\end{figure}

We report the total runtime for each task in~\cref{tab:computational-time}, measured on a single NVIDIA A100 40GB GPU. To analyze the computational overhead introduced by our framework, we measure the total time spent computing the guidance field $\Delta(x_t)$ over one generation pass with $T=50$ timesteps and batch size $N=16$, during the optimization process while the dataset size $N$ expanding. Note that all reported time measurements exclude the time required for reward evaluations. \cref{fig:ablation_time_seconds} reports the absolute wall-clock overhead in seconds, and \cref{fig:ablation_time_ratio} reports the relative overhead ratio, defined as the Flow-Direct overhead divided by the wall-clock time of the (unguided) pre-trained flow model generation. The memory consumption is shown in~\cref{fig:ablation_time_memory}.

Both runtime and memory scale linearly\footnote{The small spike at the $N=16$ is likely due to one-time GPU memory allocation during the first function calls.} with the dataset size $N$. At $N=1600$, Flow-Direct incurs approximately 10 seconds of computational wall-clock time. This introduces roughly a 30\% time overhead compared to the unguided pre-trained model's generation, while consuming 400 MB of memory.

\begin{table}[ht]
  \caption{Total runtime for each task, measured on a single NVIDIA A100 40GB GPU.}
  \label{tab:computational-time}
  \centering
  \begin{tabular}{ll}
    \toprule
    Task                 & Runtime    \\
    \midrule
    Aesthetic            & 1h 31m     \\
    HPS-v3               & 1h 18m     \\
    Compress             & 1h 9m      \\
    Incompress           & 1h 12m     \\
    Attribute Alignment  & 1h 14m     \\
    Aerodynamic          & 5h 33m     \\
    \bottomrule
  \end{tabular}
\end{table}

\section{Best-of-N Experiment}
\label[appendix]{app:best_of_n}

In this section, we ablate whether Flow-Direct can produce better samples than the pre-trained distribution. We follow the same experiment procedure in Table~\ref{tab:result}, and implement Best-of-N, which independently samples $N=1600$ samples from the pre-trained model, and report the best reward in~\cref{tab:best_of_n}. For comparison, we report the mean and best reward achieved by Flow-Direct in~\cref{tab:best_of_n}.

\begin{table}[ht]
  \caption{Best-of-N under a matched reward-evaluation budget (1600 evaluations). \textit{Best-of-N} reports the maximum reward over 1600 independently generated samples. \textit{Flow-Direct (final-batch mean)} is the average reward over the final batch of 16 samples, as reported in~\cref{tab:result}. \textit{Flow-Direct (best)} reports the maximum reward over its 1600 accumulated samples. Note: As \textit{Compressibility} is a minimization task, its values are inverted.}
  \label{tab:best_of_n}
  \centering
  \resizebox{\textwidth}{!}{%
  \begin{tabular}{lcccccccccc}
    \toprule
    \multirow{2}{*}{Algorithm} & \multirow{2}{*}{Aesthetic} & \multirow{2}{*}{HPSv3} & \multirow{2}{*}{Compressibility} & \multirow{2}{*}{Incompressibility} & \multicolumn{6}{c}{Attribute Alignment} \\
    \cmidrule(lr){6-11}
     & & & & & Happy & Fluffy & Running & Cute & Vivid & Fierce \\
    \midrule
    Best-of-N                      & $6.76{\pm}.28$ & $10.27{\pm}1.34$ & $-34.88{\pm}12.73$ & $170.93{\pm}25.51$ & $0.36{\pm}.03$ & $0.28{\pm}.01$ & $0.13{\pm}.00$ & $0.53{\pm}.00$ & $0.23{\pm}.03$ & $0.26{\pm}.00$ \\
    Flow-Direct (final-batch mean) & $7.18{\pm}.27$ & $10.94{\pm}1.06$ & $-14.63{\pm}7.99$  & $284.92{\pm}28.09$ & $0.41{\pm}.02$ & $0.42{\pm}.02$ & $0.32{\pm}.05$ & $0.52{\pm}.07$ & $0.25{\pm}.04$ & $0.30{\pm}.01$ \\
    Flow-Direct (best)             & $7.45{\pm}.25$ & $11.70{\pm}1.19$ & $-10.81{\pm}6.51$  & $298.48{\pm}27.95$ & $0.50{\pm}.03$ & $0.51{\pm}.02$ & $0.43{\pm}.06$ & $0.64{\pm}.06$ & $0.30{\pm}.04$ & $0.37{\pm}.00$ \\
    \bottomrule
  \end{tabular}%
  }
\end{table}

\cref{tab:best_of_n} shows that Flow-Direct's average performance already matches or exceeds Best-of-N's \textit{best} sample. The final-batch mean (row 2) surpasses Best-of-N (row 1) on nine of ten rewards, and is statistically indistinguishable on the remaining one (Cute: $0.521{\pm}.065$ vs. $0.527{\pm}.000$). This is the stronger comparison because it is an unselected average over 16 samples in the final batch, whereas the Best-of-N is a maximum over 1600. This indicates that Flow-Direct optimizes the generative distribution, rather than merely retrieving the best sample from the accumulated dataset.

Moreover, comparing the two maxima over 1600 samples (rows 1 and 3), Flow-Direct achieves a consistently higher reward on all ten reward functions. This is because independent sampling is limited by the support of the pre-trained distribution: reranking can only retrieve the best sample the base model happens to produce, whereas Flow-Direct progressively transports the sampling distribution toward the high-reward region.

Finally, we note that Best-of-N returns a single retained sample, while Flow-Direct returns a persistent guidance field, from which novel high-reward samples can be drawn without any further reward evaluation.

\section{Related Works}
\label[appendix]{app:related_works}

In this section, we review training-free guidance methods that support black-box rewards. These methods guide the flow model generation process by using the scalar feedback from a non-differentiable reward function. Existing approaches differ in how they use reward feedback: some approximate a local update to the denoising distribution, some maintain multiple candidate trajectories and reallocate computation based on intermediate reward estimates, and others formulate generation as an explicit search over the denoising trajectory.

\paragraph{Evolvable.}
Evolvable~\citep{wei2025evolvable} treats each denoising transition as a local black-box optimization problem. At an intermediate state, it samples multiple candidates from the current denoising distribution, queries the reward function on these candidates, and uses the resulting scalar rewards to update the mean of the next-step distribution toward higher-reward regions. This provides a derivative-free analogue of gradient guidance, since the update is estimated from reward-weighted perturbations rather than from backpropagation through the reward function.

\paragraph{FK-Steering.}
FK-Steering~\citep{singhal2025general} adopts an SMC-based particle filtering procedure to steer generation toward a reward-tilted target distribution. It maintains a population of denoising trajectories as particles, evaluates their utility using the reward function, and performs resampling so that particles with higher reward are duplicated while lower-reward particles are discarded. This particle filtering view makes the method compatible with arbitrary black-box rewards, since guidance only requires scalar reward value rather than gradients.

\paragraph{TreeG and DSearch.}
TreeG~\citep{guo2025training} formulates training-free guidance as a tree search problem over the iterative generation process. At each timestep, it selects the high-reward nodes to branch out, while pruning low-reward nodes. This effectively steers the generation trajectory toward high-reward terminal sample. Because this process relies solely on scalar feedback, it natively supports black-box reward functions. DSearch~\citep{li2025dynamic} builds on the same framework and dynamically allocates the search budget across denoising steps. It maintains more active nodes in early denoising steps to preserve trajectory diversity, and gradually reduces the number of active nodes while increasing the branching factor in later steps. This leverages the coarse-to-fine nature of the diffusion process to allocate reward evaluations more efficiently across denoising steps.

\paragraph{SDE for Flow Model.}
The methods discussed above rely on stochastic sampling dynamics, either to approximate local guidance directions or to explore multiple candidate generation trajectories. To apply them to flow models, we use the stochastic flow sampler from~\citep{liu2025flow,singh2024stochastic}, whose discretization step is given by:
\begin{align}
x_{t+\Delta t}
&= \underbrace{a_t x_t + b_t v_\theta(x_t,t) + c_t \epsilon_t}_{\texttt{step}(x_t,v_t)},
\label{eq:sde_step}
\end{align}
for vector field $v_t = v_\theta(x_t,t)$, noise $\epsilon_t \sim \mathcal{N}(0,I)$, step size $\Delta t$, and the coefficients are given by:
\begin{align}
a_t &= 1-\frac{\sigma_t^2 \Delta t}{2(1-t)}, \qquad
b_t = \Delta t\left(1+\frac{t\sigma_t^2}{2(1-t)}\right), \qquad
c_t = \sigma_t\sqrt{\Delta t}.
\end{align}
Following Flow-GRPO~\citep{liu2025flow}, we set
$\sigma_t = \eta \sqrt{(1-t)/t}$ with noise level $\eta=0.7$ in all experiments. This SDE sampler preserves the same marginal probability path as the deterministic flow ODE. When $\eta=0$, the update reduces to the deterministic ODE discretization in~\cref{eq:ode_step}.

\section{Implementation Details for~\cref{sec:3.1} Demo}
\label[appendix]{app:demo_impl}

This section provides the implementation to produce the~\cref{fig:data_guidance}.

\paragraph{Data Preparation.}
We consider three guidance tasks, where the goal is to steer the flow model toward a specific target attribute: \textit{wings}, \textit{sketch}, or \textit{cartoon}. For each task, we construct a pair of base and target datasets, each containing $64$ generated samples. All datasets are generated using the same pre-trained flow model, and the guidance computation is performed in latent space. The prompts used to construct these datasets are summarized in~\cref{tab:demo_data}.

\newcommand{\lightrule}{\specialrule{0.3pt}{0.25em}{0.25em}}

\begin{table}[h]
\centering
\caption{Base and target datasets used in the dataset-guidance demonstration. Each base and target dataset contains $64$ generated latent samples.}
\label{tab:demo_data}
{
\renewcommand{\arraystretch}{1.5}
\setlength{\tabcolsep}{4pt}
\setlength{\belowrulesep}{0pt}
\begin{tabular}{ll}
\toprule
Task & Data pair \\
\midrule
Wings
&
\begin{tabular}[c]{@{}l@{:\hspace{1em}}l@{}}
$\mathcal{D}^{\text{base}}_{\text{wings}}$
& \texttt{a puppy on grass field.} \\
$\mathcal{D}^{\text{target}}_{\text{wings}}$
& \texttt{a puppy with golden wings on grass field.}
\end{tabular}
\\[0.3em]
\lightrule

Sketch
&
\begin{tabular}[c]{@{}l@{:\hspace{1em}}l@{}}
$\mathcal{D}^{\text{base}}_{\text{sketch}}$
& \texttt{a realistic photo of a puppy.} \\
$\mathcal{D}^{\text{target}}_{\text{sketch}}$
& \texttt{a sketch drawing of a puppy.}
\end{tabular}
\\[0.3em]
\lightrule

Cartoon
&
\begin{tabular}[c]{@{}l@{:\hspace{0.7em}}l@{}}
$\mathcal{D}^{\text{base}}_{\text{cartoon}}$
& \texttt{a realistic photo of a puppy.} \\
$\mathcal{D}^{\text{target}}_{\text{cartoon}}$
& \texttt{a cartoon puppy.}
\end{tabular}
\\
\bottomrule
\end{tabular}
}
\end{table}

\paragraph{Guidance Field.} For each target, we compute the term according to~\cref{eq:data_field}. We also test composite target \textit{sketch + cartoon} by summing the guidance field. In the demo, we additionally scale each guidance field by a scale $\alpha$ to make the visual effect more apparent. This scale is introduced only for the qualitative demonstration in this section and is not used in the rest of the paper. The concrete formulation is stated in~\cref{tab:demo_guidance}.

\begin{table}[ht]
\centering
\caption{Guidance field implementation for each task.}
\label{tab:demo_guidance}
{
\renewcommand{\arraystretch}{1.5}
\setlength{\belowrulesep}{0pt}
\begin{tabular}{lll}
\toprule
Target & Guidance field & $\alpha$ \\
\midrule
Wings 
& $\Delta_{\text{wings}}(x_t) \coloneqq \alpha\,\Delta(x_t;\mathcal{D}^{\text{base}}_{\text{wings}}, \mathcal{D}^{\text{target}}_{\text{wings}})$
& $3$ \\

Sketch 
& $\Delta_{\text{sketch}}(x_t) \coloneqq \alpha\,\Delta(x_t;\mathcal{D}^{\text{base}}_{\text{sketch}}, \mathcal{D}^{\text{target}}_{\text{sketch}})$
& $2$ \\

Cartoon 
& $\Delta_{\text{cartoon}}(x_t) \coloneqq \alpha\,\Delta(x_t;\mathcal{D}^{\text{base}}_{\text{cartoon}}, \mathcal{D}^{\text{target}}_{\text{cartoon}})$
& $2$ \\

Sketch + Cartoon 
& $\Delta_{\text{sketch+cartoon}}(x_t) \coloneqq \alpha\left(\Delta_{\text{sketch}}(x_t) + \Delta_{\text{cartoon}}(x_t)\right)$
& $1$ \\
\bottomrule
\end{tabular}%
}
\end{table}

\paragraph{Generation.}
We evaluate the generation procedure using two inference prompts:
\begin{center}
\texttt{a puppy on grass field.} \qquad
\texttt{a cat sit on grass field.}
\end{center}
The first \textit{puppy} prompt matches the dataset construction setting, whereas the second \textit{cat} prompt creates a distribution mismatch with the guidance datasets. The generated images are reported in~\cref{fig:data_guidance}.

\section{Implementation for Attribute Alignment}
\label[appendix]{app:attribute_alignment}

In this section, we describe the implementation details for the \textit{attribute alignment} reward. The goal of this reward is to assess whether a generated image aligns with a specified semantic attribute.

We evaluate the generated image using an open-source vision-language model (VLM). Specifically, we use Gemma 4~\citep{team2024gemma} (\texttt{google/gemma-4-E4B-it}). We format the image and text as a joint multimodal input message, and ask the VLM the following question:
\begin{quote}
\texttt{Is the \{prompt\} \{attribute\}? Strictly answer "yes" or "no" without any explanation.}
\end{quote}
The placeholder \texttt{\{prompt\}} represents the subject and \texttt{\{attribute\}} is the target semantic attribute.

Rather than decoding the VLM's output into a binary reward value, we implement Soft-TIFA~\citep{kamath2025geneval} for a soft reward value. Specifically, Soft-TIFA extracts the VLM's response logits and computes the probability of the "\texttt{yes}" token. This probability then serves as a continuous reward within range $[0, 1]$, effectively quantifying the model's confidence that the image matches the target attribute.

In our experiments, each animal prompt is paired with one predefined target attribute, as stated in~\cref{tab:attribute_alignment_pairs}. During optimization, the flow model is optimized to generate images that match the corresponding target attribute.

\begin{table}[ht]
\centering
\caption{Target attribute used in our experiment.}
\begin{tabular}{ll}
\toprule
Animal prompt & Target attribute \\
\midrule
\texttt{dog}   & \texttt{happy} \\
\texttt{cat}   & \texttt{fluffy} \\
\texttt{horse} & \texttt{running} \\
\texttt{bear}  & \texttt{cute} \\
\texttt{bird}  & \texttt{vivid} \\
\texttt{lion}  & \texttt{fierce} \\
\bottomrule
\end{tabular}
\label{tab:attribute_alignment_pairs}
\end{table}

\section{Implementation for Prompt Alignment}
\label[appendix]{app:prompt_alignment}

In this section, we describe the implementation details for the \textit{prompt alignment} reward. The goal is to assess whether a generated image aligns with the specific complex prompt. This is a genuinely expensive reward, since each evaluation requires a paid LLM API query, so query efficiency becomes more important than memory or computational efficiency.

Specifically, we align the flow model to the following challenging prompt that involves object counting and spatial arrangement:
\begin{quote}
\texttt{Seven apples arranged in a perfect circle, each in a distinct rainbow color (red, orange, yellow, green, blue, indigo, and violet), and a single transparent glass apple in the center.}
\end{quote}

Each generated image is evaluated by Gemini~\citep{reid2024gemini,comanici2025gemini} (\texttt{gemini-3.6-flash}) at a cost of \$0.002611 (USD) per query. The actual query to the LLM is:
\begin{quote}
\texttt{Rate how accurately this image matches the text prompt: "\{prompt\}"}\\
\texttt{Assign a rating from 0 to 10, where 0 means does not match at all, and 10 means perfect match.}
\end{quote}
The placeholder \texttt{\{prompt\}} represents the target text prompt. We constrain the LLM to respond with a structured output containing a single numerical rating, and use this rating directly as the reward value within range $[0, 10]$. We set the decoding temperature to $0$, with the thinking level and the image resolution both set to \texttt{low}.

We compare all methods under matched query budgets of $200$, $400$, and $800$ evaluations, repeat each experiment over three seeds, and report the average rating over the final batch of generated samples for each method in~\cref{tab:prompt_alignment}.

\begin{table}[ht]
  \caption{Average rating under a matched LLM API query budget, with the corresponding cost. For each run we record the average rating over the final batch of generated samples, and report the mean and standard deviation across three seeds.}
  \label{tab:prompt_alignment}
  \centering
  \begin{tabular}{lccc}
    \toprule
    Algorithm   & 200 (\$0.52) & 400 (\$1.04) & 800 (\$2.09) \\
    \midrule
    Evolvable   & $1.8{\pm}2.8$ & $1.2{\pm}2.0$ & $2.0{\pm}2.6$ \\
    FK-Steering & $5.6{\pm}2.7$ & $5.9{\pm}1.5$ & $6.9{\pm}.7$ \\
    TreeG       & $4.9{\pm}1.5$ & $4.9{\pm}1.2$ & $5.0{\pm}.4$ \\
    DSearch     & $\mathbf{7.7}{\pm}1.7$ & $6.8{\pm}.9$ & $7.5{\pm}.8$ \\
    Flow-Direct & $6.8{\pm}.3$ & $\mathbf{7.9}{\pm}.4$ & $\mathbf{9.0}{\pm}.5$ \\
    \bottomrule
  \end{tabular}
\end{table}

\section{Implementation Details for 3D Vehicle Optimization}
\label[appendix]{app:domino}

\subsection{3D Generation Model}
For the 3D vehicle generation task, we utilize the TRELLIS~\citep{xiang2025structured} text-to-3D model (\texttt{microsoft/TRELLIS-text-xlarge}). The TRELLIS pipeline consists of two sequential stages: 1) generating a sparse structure, and 2) decoding and refining this sparse structure into a dense, high-quality 3D mesh. In our experiments, we apply the guidance \textit{exclusively} to the first stage. We fix the generation prompt to \textit{car}, with the goal of guiding the first stage to generate highly aerodynamic cars.

\subsection{DoMINO Reward Evaluation}
To evaluate the aerodynamic performance of the generated cars, we utilize the DoMINO predictive model~\citep{ranade2025domino} (\texttt{nvcr.io/nim/nvidia/domino-automotive-aero:2.1.0-41313772}).

The primary optimization objective for this task is the \textit{drag coefficient} ($C_d$). While the DoMINO API directly outputs the raw drag force ($F_d$) evaluated on the vehicle mesh, we calculate the dimensionless drag coefficient using the standard aerodynamic equation:
$$C_d = \frac{F_d}{\frac{1}{2} \rho V^2 A}$$
where:
\begin{itemize}
    \item $F_d$ is the drag force in Newtons returned by the DoMINO model.
    \item $A$ is the frontal area of the vehicle mesh (in square meters), calculated by projecting the silhouette of the generated vehicle onto the YZ-plane.
    \item $\rho$ is the standard air density at sea level, set to a constant $\rho = 1.225 \text{ kg/m}^3$.
    \item $V$ is the freestream velocity, set to $V = 30.0 \text{ m/s}$.
\end{itemize}

\section{Additional Qualitative Results}
\label[appendix]{app:additional_qualitative}

\subsection{3D Vehicle Aerodynamic Optimization}
\begin{figure}[htbp]
    \centering
    % Left Subfigure
    \begin{subfigure}[b]{0.45\textwidth}
        \centering
        \includegraphics[width=\textwidth]{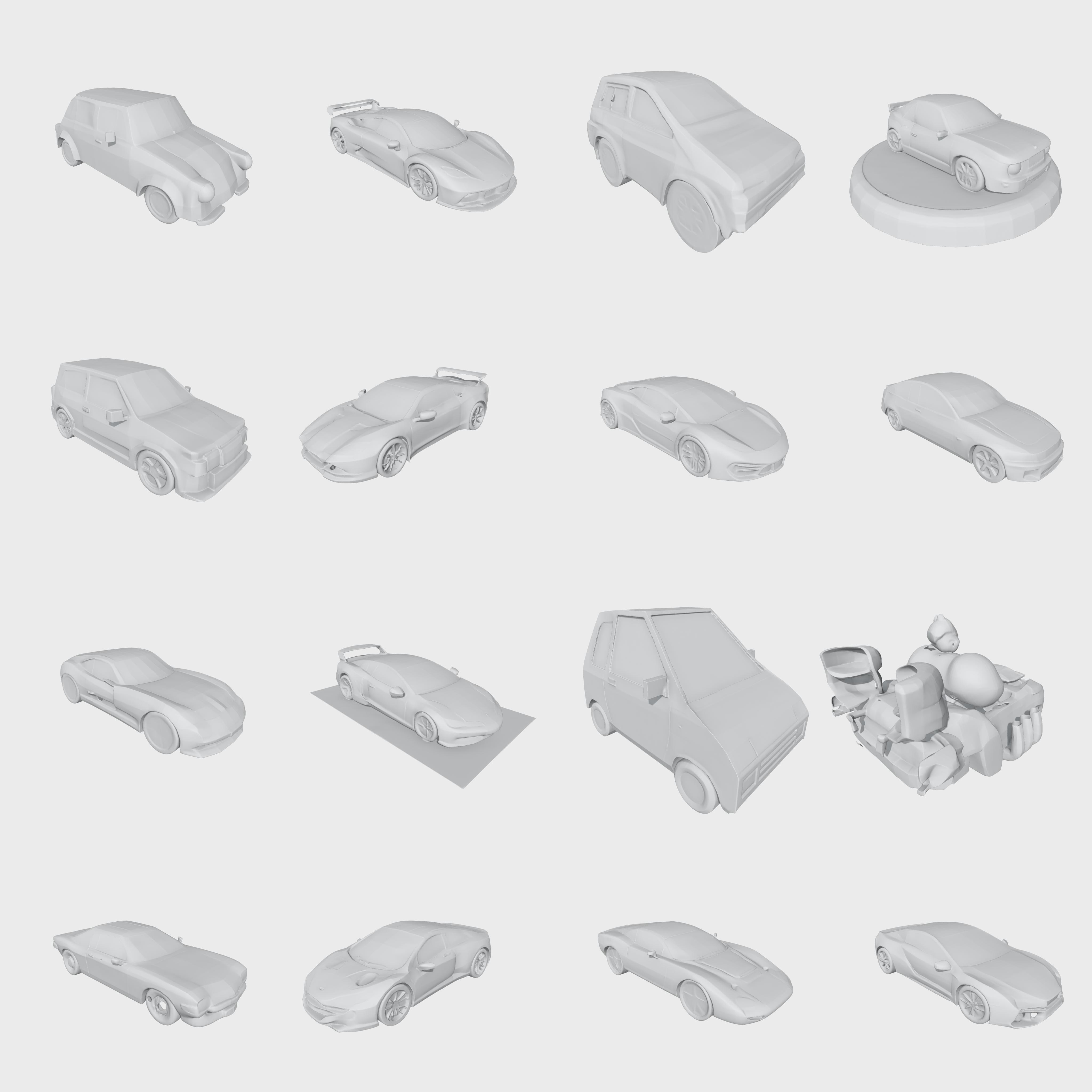}
        \caption{Unguided}
        \label{fig:car-before}
    \end{subfigure}
    \hfill
    \begin{subfigure}[b]{0.45\textwidth}
        \centering
        \includegraphics[width=\textwidth]{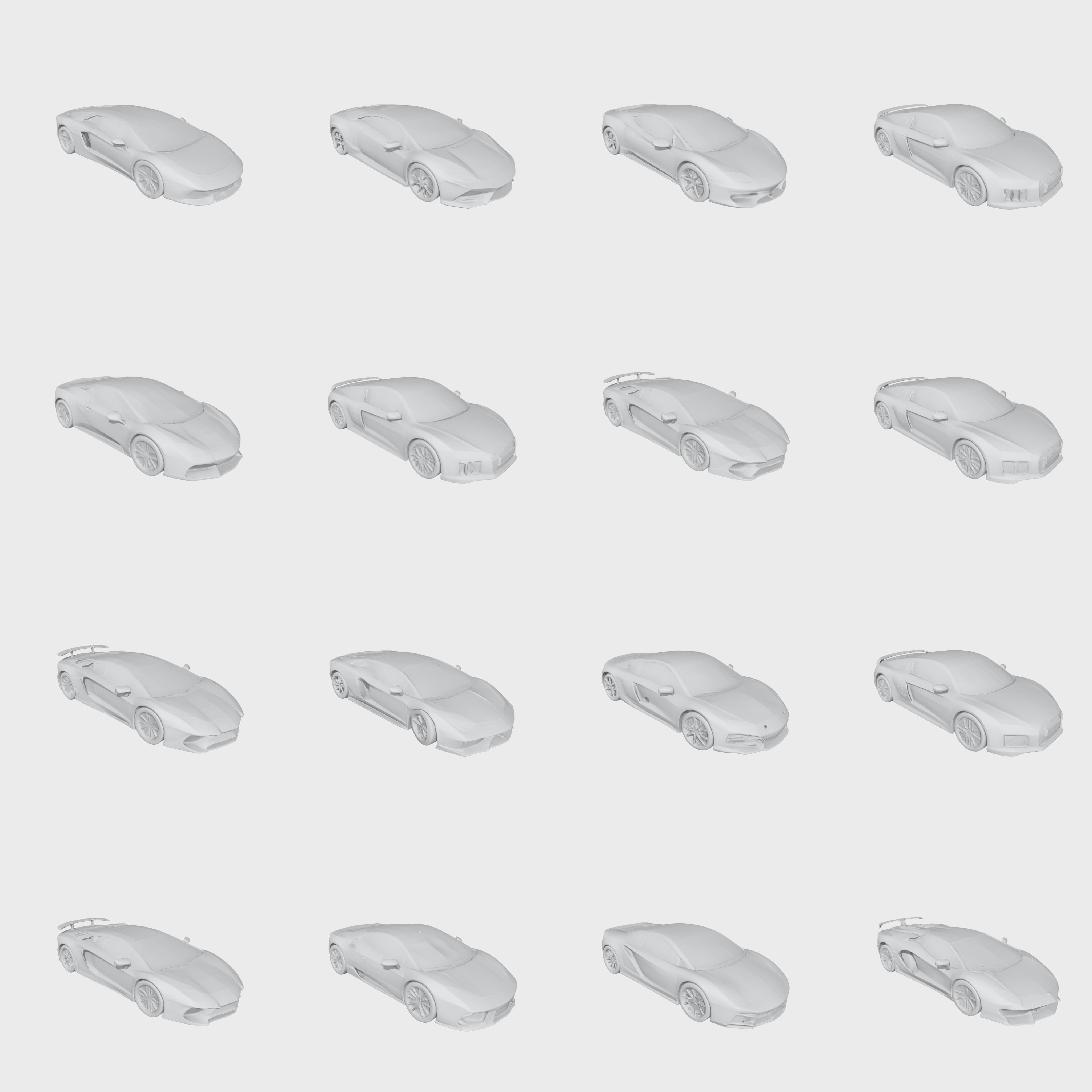}
        \caption{Flow-Direct}
        \label{fig:car-after}
    \end{subfigure}
    \caption{Qualitative results for 3D vehicle aerodynamic optimization.}
    \label{fig:car-comparison}
\end{figure}

\label[appendix]{app:additional_qualitative_3d}
In this section, we provide additional qualitative results for the 3D vehicle aerodynamic optimization task. In \cref{fig:car-comparison}, we compare 16 randomly generated 3D vehicle meshes produced by the unguided pre-trained TRELLIS model with those optimized by Flow-Direct. We observe that the optimized car structures exhibit highly aerodynamic shapes while preserving their overall structural integrity.

\subsection{Image Alignment}
\label[appendix]{app:additional_qualitative_image}

In this section, we provide additional qualitative results for all reward functions considered in the main experiments. Each figure shows samples generated by different inference-time guidance methods under the same reward-evaluation budget. Columns correspond to prompts or target attributes, and rows correspond to the optimization method. These results complement~\cref{fig:result} and further illustrate that Flow-Direct consistently produces samples that better reflect the target reward while preserving image quality.

\begin{figure}[ht]
  \centering
  \includegraphics[width=0.8\linewidth]{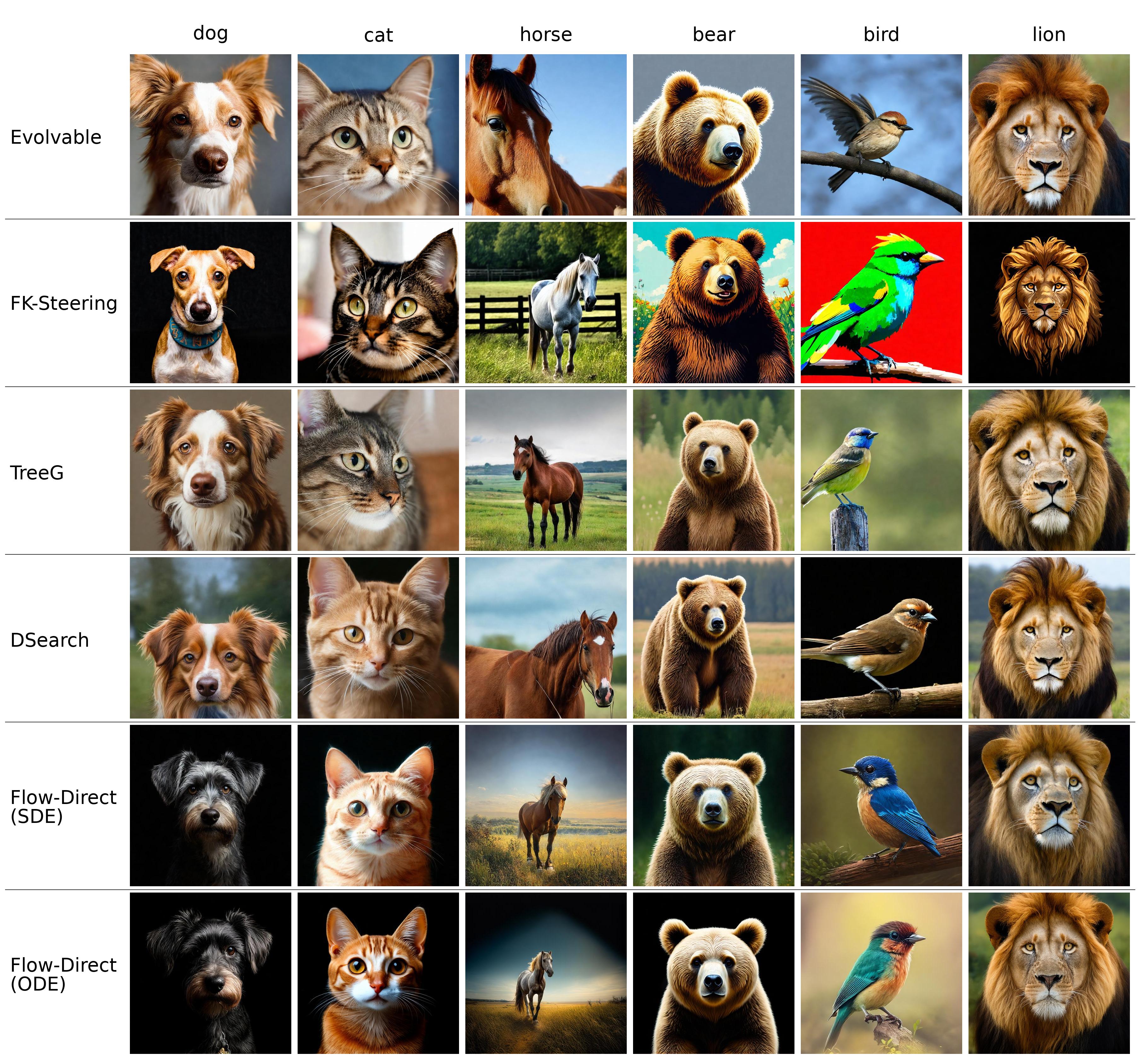}
  \caption{Additional qualitative results for the \textit{Aesthetic} reward.}
  \label{fig:app_aesthetic}
\end{figure}

\begin{figure}[ht]
  \centering
  \includegraphics[width=0.8\linewidth]{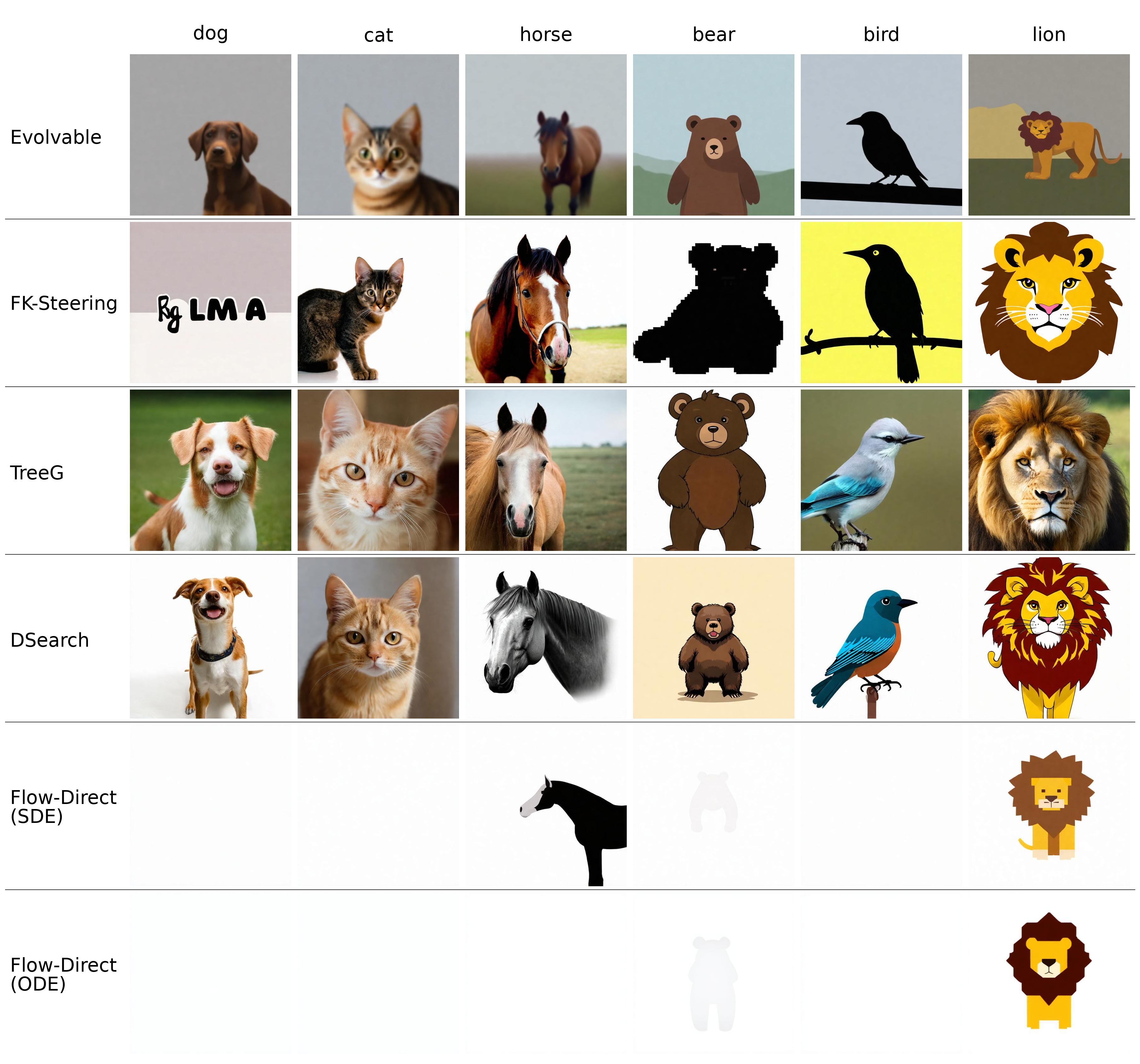}
  \caption{Additional qualitative results for the \textit{Compressibility} reward.}
  \label{fig:app_compress}
\end{figure}

\begin{figure}[ht]
  \centering
  \includegraphics[width=0.8\linewidth]{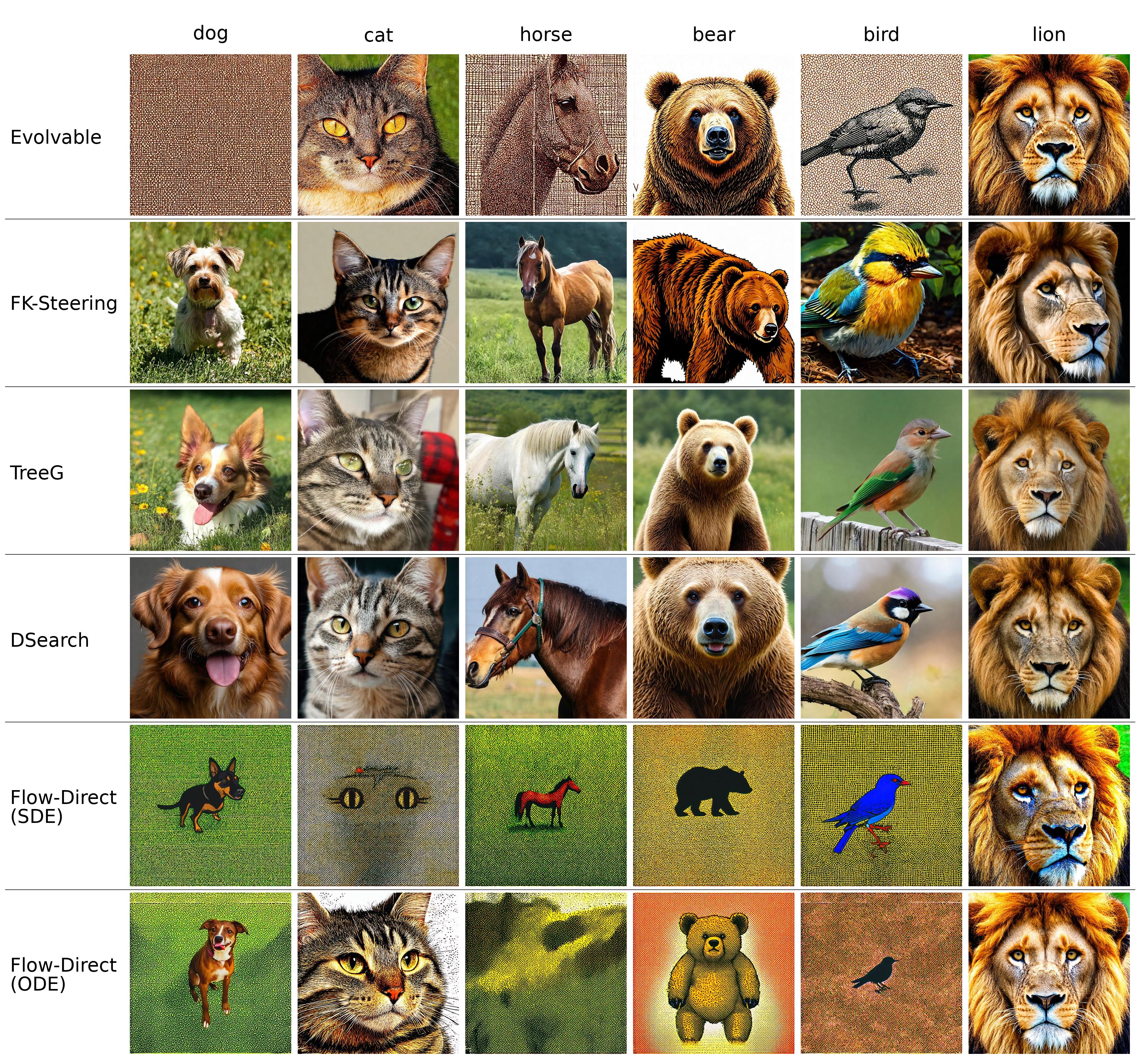}
  \caption{Additional qualitative results for the \textit{Incompressibility} reward.}
  \label{fig:app_incompress}
\end{figure}

\begin{figure}[ht]
  \centering
  \includegraphics[width=0.8\linewidth]{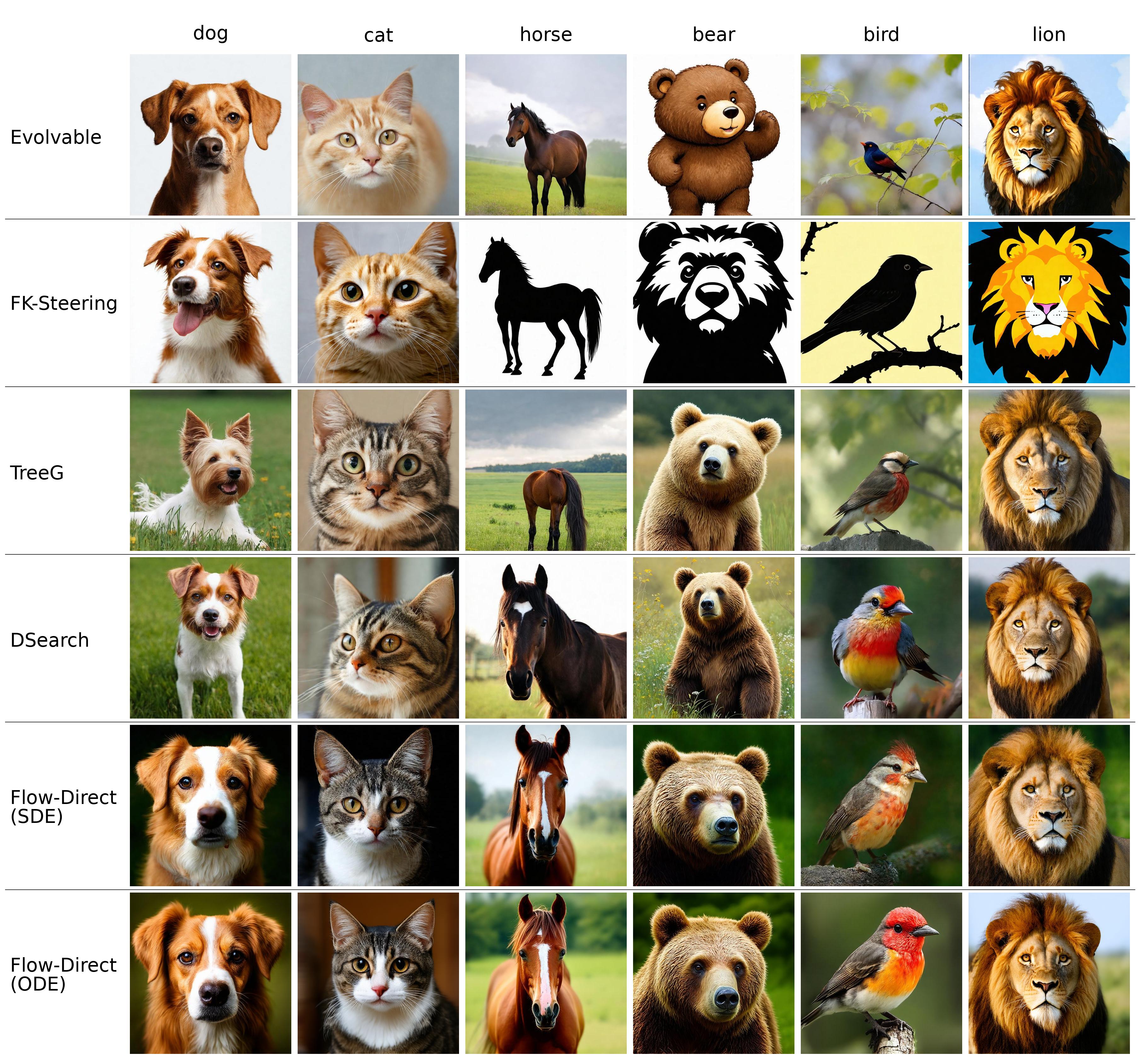}
  \caption{Additional qualitative results for the \textit{HPSv3} reward.}
  \label{fig:app_hps}
\end{figure}

\begin{figure}[ht]
  \centering
  \includegraphics[width=0.8\linewidth]{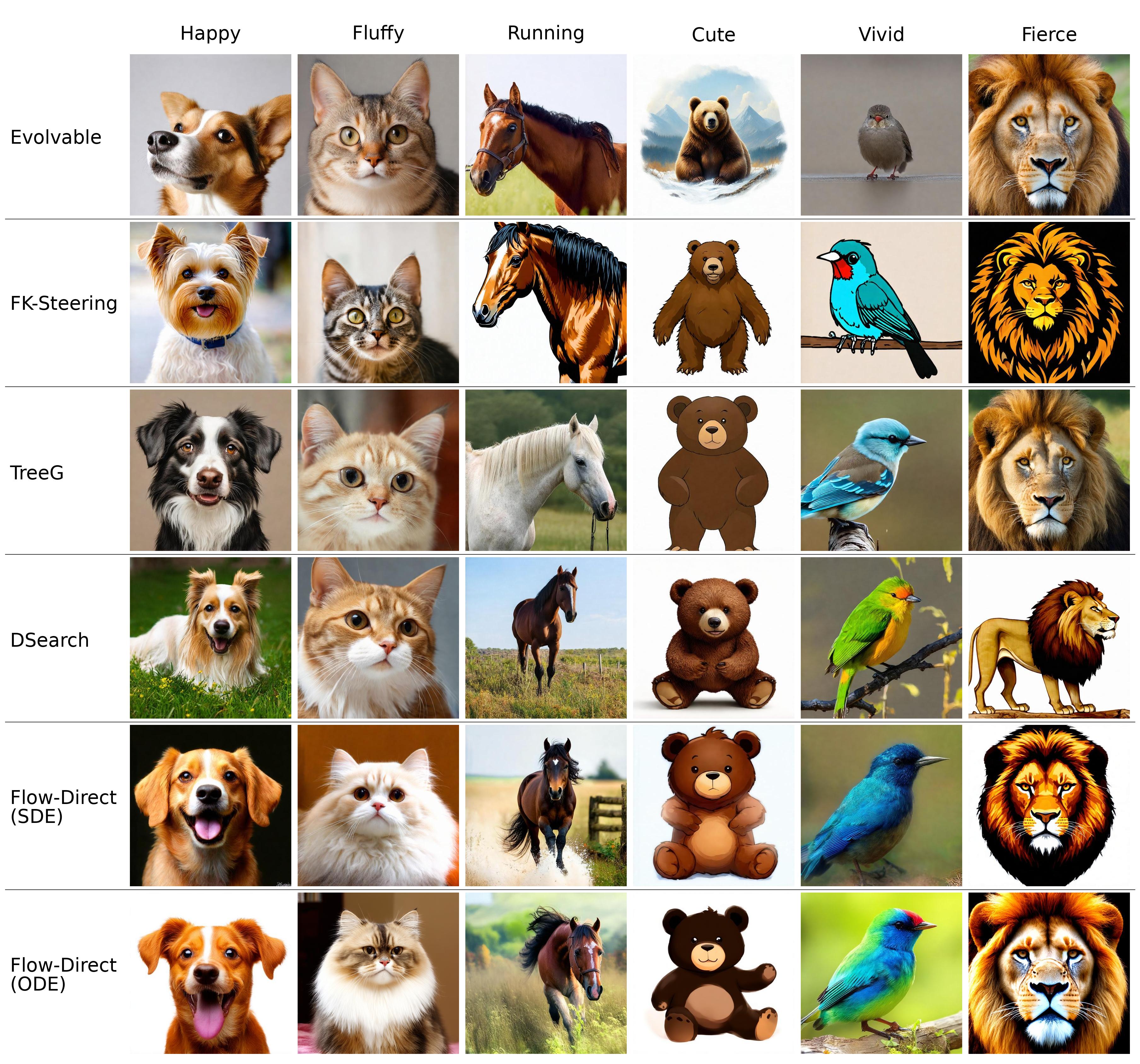}
  \caption{Additional qualitative results for the \textit{Attribute Alignment} reward.}
  \label{fig:app_gemma}
\end{figure}

%%%%%%%%%%%%%%%%%%%%%%%%%%%%%%%%%%%%%%%%%%%%%%%%%%%%%%%%%%%%

\clearpage
\newpage
\clearpage
\section*{NeurIPS Paper Checklist}

\begin{enumerate}

\item {\bf Claims}
    \item[] Question: Do the main claims made in the abstract and introduction accurately reflect the paper's contributions and scope?
    \item[] Answer: \answerYes{} % Replace by \answerYes{}, \answerNo{}, or \answerNA{}.
    \item[] Justification: We make three claims: (1) a guidance field formulation, (2) feedback-efficiency, (3) reusability. Claim (1) is justified theoretically in Section 3 and Appendix A; claims (2) and (3) are justified empirically by the experiments in Sections 4.1 and 4.2 respectively.

\item {\bf Limitations}
    \item[] Question: Does the paper discuss the limitations of the work performed by the authors?
    \item[] Answer: \answerYes{} % Replace by \answerYes{}, \answerNo{}, or \answerNA{}.
    \item[] Justification: Section 5 specifically discusses the limitations.

\item {\bf Theory assumptions and proofs}
    \item[] Question: For each theoretical result, does the paper provide the full set of assumptions and a complete (and correct) proof?
    \item[] Answer: \answerYes{} % Replace by \answerYes{}, \answerNo{}, or \answerNA{}.
    \item[] Justification: The two main theoretical results, Propositions 1 and 2, are stated formally with assumptions in Section 3, with the complete derivations are provided in Appendix A.

    \item {\bf Experimental result reproducibility}
    \item[] Question: Does the paper fully disclose all the information needed to reproduce the main experimental results of the paper to the extent that it affects the main claims and/or conclusions of the paper (regardless of whether the code and data are provided or not)?
    \item[] Answer: \answerYes{} % Replace by \answerYes{}, \answerNo{}, or \answerNA{}.
    \item[] Justification: Algorithm 1 fully specifies the proposed method, with the practical implementation details provided in~\cref{app:implementation}. Other experimental details are provided in~\cref{app:demo_impl,app:attribute_alignment,app:prompt_alignment,app:domino}.

\item {\bf Open access to data and code}
    \item[] Question: Does the paper provide open access to the data and code, with sufficient instructions to faithfully reproduce the main experimental results, as described in supplemental material?
    \item[] Answer: \answerYes{} % Replace by \answerYes{}, \answerNo{}, or \answerNA{}.
    \item[] Justification: The source code is publicly available at: \url{https://github.com/kimyong95/finetune-stable-diffusion}.

\item {\bf Experimental setting/details}
    \item[] Question: Does the paper specify all the training and test details (e.g., data splits, hyperparameters, how they were chosen, type of optimizer) necessary to understand the results?
    \item[] Answer: \answerYes{} % Replace by \answerYes{}, \answerNo{}, or \answerNA{}.
    \item[] Justification: Section 4.1 specifies all the hyperparameters.

\item {\bf Experiment statistical significance}
    \item[] Question: Does the paper report error bars suitably and correctly defined or other appropriate information about the statistical significance of the experiments?
    \item[] Answer: \answerYes{} % Replace by \answerYes{}, \answerNo{}, or \answerNA{}.
    \item[] Justification: Tables 1 and 3 report mean reward value ± standard deviation across multiple runs.

\item {\bf Experiments compute resources}
    \item[] Question: For each experiment, does the paper provide sufficient information on the computer resources (type of compute workers, memory, time of execution) needed to reproduce the experiments?
    \item[] Answer: \answerYes{} % Replace by \answerYes{}, \answerNo{}, or \answerNA{}.
    \item[] Justification: The compute resources are stated in Appendix C.
    
\item {\bf Code of ethics}
    \item[] Question: Does the research conducted in the paper conform, in every respect, with the NeurIPS Code of Ethics \url{https://neurips.cc/public/EthicsGuidelines}?
    \item[] Answer: \answerYes{} % Replace by \answerYes{}, \answerNo{}, or \answerNA{}.
    \item[] Justification: The research conforms with the NeurIPS Code of Ethics: it does not involve human subjects, the pre-trained generative models and reward predictors are publicly available academic releases, and no personally identifiable data is used.

\item {\bf Broader impacts}
    \item[] Question: Does the paper discuss both potential positive societal impacts and negative societal impacts of the work performed?
    \item[] Answer: \answerYes{} % Replace by \answerYes{}, \answerNo{}, or \answerNA{}.
    \item[] Justification: Section 5 specifically discusses both positive and negative broader impacts.
    
\item {\bf Safeguards}
    \item[] Question: Does the paper describe safeguards that have been put in place for responsible release of data or models that have a high risk for misuse (e.g., pre-trained language models, image generators, or scraped datasets)?
    \item[] Answer: \answerNA{} % Replace by \answerYes{}, \answerNo{}, or \answerNA{}.
    \item[] Justification: Flow-Direct operates on top of open source models during inference-time, we do not release any new data or models, so no additional release safeguards are required.

\item {\bf Licenses for existing assets}
    \item[] Question: Are the creators or original owners of assets (e.g., code, data, models), used in the paper, properly credited and are the license and terms of use explicitly mentioned and properly respected?
    \item[] Answer: \answerYes{} % Replace by \answerYes{}, \answerNo{}, or \answerNA{}.
    \item[] Justification: All pre-trained models and baseline methods used in the paper are properly cited, and we use them in accordance with their respective public release licenses and intended research use.

\item {\bf New assets}
    \item[] Question: Are new assets introduced in the paper well documented and is the documentation provided alongside the assets?
    \item[] Answer: \answerNA{} % Replace by \answerYes{}, \answerNo{}, or \answerNA{}.
    \item[] Justification: The paper is algorithmic, and does not release any new datasets or pre-trained models.

\item {\bf Crowdsourcing and research with human subjects}
    \item[] Question: For crowdsourcing experiments and research with human subjects, does the paper include the full text of instructions given to participants and screenshots, if applicable, as well as details about compensation (if any)? 
    \item[] Answer: \answerNA{} % Replace by \answerYes{}, \answerNo{}, or \answerNA{}.
    \item[] Justification: The paper does not involve any crowdsourcing or human subjects.

\item {\bf Institutional review board (IRB) approvals or equivalent for research with human subjects}
    \item[] Question: Does the paper describe potential risks incurred by study participants, whether such risks were disclosed to the subjects, and whether Institutional Review Board (IRB) approvals (or an equivalent approval/review based on the requirements of your country or institution) were obtained?
    \item[] Answer: \answerNA{} % Replace by \answerYes{}, \answerNo{}, or \answerNA{}.
    \item[] Justification: The paper does not involve any human subjects.

\item {\bf Declaration of LLM usage}
    \item[] Question: Does the paper describe the usage of LLMs if it is an important, original, or non-standard component of the core methods in this research? Note that if the LLM is used only for writing, editing, or formatting purposes and does \emph{not} impact the core methodology, scientific rigor, or originality of the research, declaration is not required.
    %this research? 
    \item[] Answer: \answerNo{} % Replace by \answerYes{}, \answerNo{}, or \answerNA{}.
    \item[] Justification: LLM is not as a component of the proposed method.

\end{enumerate}

\end{document}